\documentclass[times,10pt]{elsarticle}

\usepackage{amssymb}
\usepackage{amsmath}

\usepackage{hyperref}
\usepackage{url}
\usepackage{graphicx}
\usepackage{booktabs}
\usepackage{multirow}
\usepackage{wrapfig}
\usepackage{xcolor}
\usepackage[ruled,vlined]{algorithm2e}

\journal{Pattern Recognition}

\begin{document}

\begin{frontmatter}



\title{Describe-to-Score: A Text-Guided Framework for Image Complexity Assessment}

\author[1]{Shipeng Liu}
\ead{lsp@xauat.edu.cn}

\author[2,3]{Liang Zhao\corref{cor1}}
\ead{zhaoliang@xauat.edu.cn}

\author[4]{Dengfeng Chen}
\ead{chdengf@xauat.edu.cn}

\author[4]{Zhonglin Zhang}
\ead{zh_zhonglin@163.com}

\cortext[cor1]{Corresponding author}

\affiliation[1]{
            department={School of Mechanical and Electrical Engineering},
            organization={Xi'an University of Architecture and Technology},
            city={Xi'an},
            postcode={710055}, 
            state={Shaanxi},
            country={China}}
\affiliation[2]{
            organization={Shaanxi Key Laboratory of Geotechnical and Underground Space Engineering},
            city={Xi'an},
            postcode={710055}, 
            state={Shaanxi},
            country={China}}
\affiliation[3]{
            department={School of Artificial Intelligence and Robotics},
            organization={Xi'an University of Architecture and Technology},
            city={Xi'an},
            postcode={710055}, 
            state={Shaanxi},
            country={China}}
\affiliation[4]{
            department={School of Building Equipment Science and Engineering},
            organization={Xi'an University of Architecture and Technology},
            city={Xi'an},
            postcode={710055}, 
            state={Shaanxi},
            country={China}}

\begin{abstract}
Accurately assessing image complexity (IC) is essential for many vision tasks, yet existing approaches rely almost exclusively on visual features and therefore fail to capture the high-level semantics that humans often use when judging complexity. We introduce a multimodal perspective for IC modeling by integrating visual representations with caption-derived textual semantics. This integration enriches the representational space and provides complementary structural cues that are difficult to infer from vision alone. From an information theoretic and representation viewpoint, we offer an idealized analysis suggesting how semantic guidance can regularize the hypothesis space and support more stable generalization. We propose D2S (\textbf{D}escribe-to-\textbf{S}core), a text-guided framework that uses caption-derived semantics only during training to regularize visual complexity modeling, while preserving a vision-only inference pipeline with no additional multimodal overhead at inference. Concretely, D2S transfers semantic structure into the visual branch through feature alignment and entropy distribution alignment, encouraging the visual encoder to internalize complexity-relevant semantic regularities. Experiments show that D2S achieves state-of-the-art performance on the IC9600 benchmark and remains competitive on no-reference image quality assessment (NR-IQA) tasks. Additional analyses further clarify the sample-imbalance issue in the small samples training setting and the distribution-shift limitations observed in cross-dataset transfer. Code is available at: \url{https://github.com/xauat-liushipeng/D2S}
\end{abstract}




\end{frontmatter}



\section{Introduction}
\label{sec1:intro}

Image complexity~\citep{forsythe2009visual} (IC) is a fundamental factor in human visual perception, influencing aesthetic judgment and memorability~\citep{singh2017review}. In computer vision, accurate image complexity assessment (ICA) supports automatic annotation, active learning, and hard example mining by identifying informative samples and improving learning efficiency and generalization~\citep{feng2022ic9600}. Early approaches relied on statistical measures such as fractal dimension, entropy~\citep{li2021evaluation}, and edge density~\citep{dai2022visual} to characterize structural richness. However, these handcrafted indicators suffer from subjective criteria, inconsistent definitions, and limited cross-domain robustness. With the development of deep learning, convolutional neural networks (CNNs) and vision transformers (ViTs)~\citep{liu2025clicv2} have significantly improved prediction accuracy by learning hierarchical visual features. Nevertheless, current models still primarily emphasize low-level cues such as texture and color, while lacking explicit mechanisms to capture high-level semantics including object count, category, and spatial organization. In addition, many visual-only models provide limited interpretability~\citep{chen2015research}. More importantly, most existing ICA methods remain purely visual during learning and therefore lack an explicit mechanism for injecting high-level semantic supervision into the representation formation process.

Recent work has begun incorporating semantic signals, such as object counts~\citep{shen2024simplicity} or motion patterns~\citep{li2025micm}, into complexity modeling and has reported promising improvements. Meanwhile, we note that human perception of complexity depends on an integrated understanding of both local visual attributes and high-level semantics, including the number, types, and relationships of objects as well as implied events. This observation motivates the following central question:

\textit{How can image complexity be computationally evaluated in a human-like manner that integrates low-level visual information with high-level semantic cues?}

\begin{wrapfigure}{r}{0.5\textwidth}
\vspace{-10pt}
\includegraphics[width=0.98\linewidth]{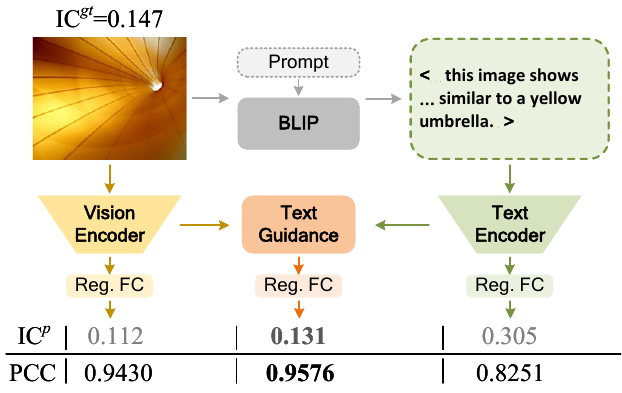}
\caption{Text-guided image complexity assessment. $\mathrm{IC}^{gt}$ and $\mathrm{IC}^{p}$ denote the ground-truth and predicted IC scores, respectively.}
\label{teaser}
\end{wrapfigure}

To explore this question, we first examined whether captions contain useful IC-related information. We used a vision–language model (VLM)~\citep{li2022blip} to generate captions for images and trained a model to predict IC based solely on textual descriptions. Surprisingly, caption-only prediction already achieved a PCC of 0.8251, as illustrated in Figure~\ref{teaser} (right) and detailed in Table~\ref{tab:7}. This result suggests that captions indeed encode meaningful semantic attributes correlated with perceived complexity. Building on this observation, we integrate the visual and textual branches in Figure~\ref{teaser} and introduce D2S (\textbf{D}escribe-to-\textbf{S}core), which leverages caption-derived semantics to guide visual complexity modeling.

The core idea of D2S is \textbf{\textit{Describe first; then Score.}}
\textbf{(1) Describing.} Captions generated by a pretrained VLM capture object concepts, relations, and structural semantics.
\textbf{(2) Scoring.} Through vision–text alignment, semantic information shapes the visual representation space, enabling the model to utilize both high-level semantics and low-level cues for more reliable complexity assessment. On IC9600~\citep{feng2022ic9600}, text-guided prediction improves PCC by 0.0146 over the visual-only baseline (Figure~\ref{teaser}, middle). Unlike conventional multimodal predictors that require both image and text inputs at inference, D2S uses language only as a training-time supervisory prior and retains a unimodal visual prediction pathway at deployment.

We further examine the rationale behind multimodal fusion~\citep{park2023visual}. From an information theoretic viewpoint, entropy analysis~\citep{yang2017feature} shows that fused visual–text features exhibit higher entropy than vision-only features, providing a richer approximation of true complexity distributions (Section~\ref{sec3.2}). From a generalization-theoretic perspective, an empirical Rademacher-complexity analysis~\citep{mohri2008rademacher} characterizes the semantic-consistency constraints as restricting the admissible family of vision-only predictors (Section~\ref{sec3.3}). Guided by these insights, we introduce entropy distribution alignment (EAL) and feature alignment (FAL) to improve cross-modal consistency. Extensive experiments validate their effectiveness: D2S achieves state-of-the-art (SOTA) performance on IC9600 with reduced inference latency, as demonstrated in Section \ref{sec-exp-setting}. Moreover, D2S shows competitive transferability to no-reference image quality assessment (NR-IQA)~\citep{mittal2012no}, including strong results on KADID-10K~\citep{lin2019kadid}. Our main contributions are as follows:

\textbullet\ We propose D2S, a text-guided framework for image complexity assessment in which language is used only during training as a semantic supervisory prior, while inference remains purely vision-based with no additional multimodal overhead.

\textbullet\ We introduce two complementary alignment mechanisms, feature alignment and entropy distribution alignment, to transfer semantic structure into the visual representation and improve cross-modal consistency during learning.

\textbullet\ We provide extensive empirical evidence showing that the proposed training-time semantic guidance improves accuracy, sample efficiency, and transfer behavior on image complexity assessment and related perceptual prediction tasks, while also clarifying its limitations under long-tailed sampling and cross-dataset distribution shift.

A more detailed discussion of related ICA and vision-language modeling studies is provided in \ref{supp-related works}.

\section{Preliminary}
\label{sec3:pre}

\subsection{Task Definition}
\label{sec3.1}
Given an input image $I$ from the dataset $\mathcal{D}$, its ground-truth complexity label is denoted as $y \in (0,1)$. An arbitrary caption generation model $g_\phi$ produces a text description $S = g_\phi(I)$. We then train a VLM $f_\theta$, which consists of a visual encoder and a text encoder, to predict the IC score $\hat{y} = f_\theta(I, S)$. The learning objective is to minimize the prediction error:
\begin{equation}
    \min_{\theta}\, \mathbb{E}_{(I,y)\sim\mathcal{D}}\big[\ell\big(f_{\theta}(I,S),y\big)\big]
\end{equation}
where $\ell$ denotes the loss function (e.g., MSE), and $\theta$ represents the model parameters. For the single-modal case using only the image input, the prediction result is written as $\hat{y}_v = f_\theta(I)$. In what follows, we focus on understanding how multimodal training reshapes the representation and why semantic information can help complexity assessment under suitable conditions. We do not assume a priori that $\hat{y}$ always dominates $\hat{y}_{v}$. Instead, we analyze the mechanisms through which multimodal fusion can provide advantages over the visual-only baseline.

\subsection{Entropy and Complexity}
\label{sec3.2}
\textbf{Proposition 1.} \textit{Image complexity tends to increase with the growth of visual diversity and semantic diversity. Let the original entropy be $H(I)$, the entropy of visual features be $H_v^F(I)$, and the entropy of semantic features be $H_s^F(S)$. We model the fused entropy as}
\begin{equation}
    H^{F}(I) = \alpha\, H_{v}^{F}(I) + \beta\, H_{s}^{F}(S)
\end{equation}
\textit{where $\alpha, \beta > 0$ are weighting coefficients.} This linear relation is introduced as a tractable approximation to capture the joint contribution of visual and semantic cues, rather than as a universal identity for all multimodal systems. We empirically verify in \ref{supp-alpha-beta} that, for our trained model, there exist positive coefficients $\alpha$ and $\beta$ such that $H^{F}(I)$ closely matches the entropy measured from fused features.

\textbf{Implications of Proposition 1.} Semantic information provides complementary cues to visual information and can increase the diversity of the learned representation when textual features capture complexity-relevant factors. In this work, feature entropy is used as a proxy for representational diversity rather than as a direct measurement of perceived image complexity. Therefore, a higher fused entropy should not be interpreted as a universal guarantee of better prediction accuracy. Instead, it indicates that multimodal training can expose the visual branch to additional semantic factors, such as object categories, object count, spatial organization, and scene context, which are often involved in human judgments of image complexity. Although $H^{F}(I) > H_v^{F}(I)$ does not hold universally for all choices of $\alpha$ and $\beta$, empirical analysis shows that the fused entropy tends to exceed visual entropy for the vast majority of samples in our trained model. This observation supports the use of entropy-based alignment as a training-time regularizer that encourages cross-modal consistency rather than as an absolute definition of image complexity.

\subsection{Semantic Constraints on the Admissible Predictor Family}
\label{sec3.3}

The entropy analysis above characterizes how caption-derived semantics may introduce complementary structural information into the learned representation. Representational richness, however, should be distinguished from the capacity of the corresponding predictor family. The intrinsic dimension of a representation describes the effective number of variation directions occupied by feature samples, whereas empirical Rademacher complexity characterizes the ability of a family of prediction functions to fit random signs on a finite sample. These two quantities describe different properties of a learning system and do not, in general, exhibit a monotonic relationship.

In D2S, semantic alignment may broaden the effective geometry of the visual representation while restricting the set of predictors that are consistent with the caption-derived semantic structure. We therefore analyze multimodal guidance from the perspective of a semantically admissible predictor family.

\textbf{Definition 1 (Empirical Rademacher Complexity).}

\textit{Given a sample set $\mathcal{D}_{n}=\{x_i\}_{i=1}^{n}$ and a hypothesis family $\mathcal{F}$, the empirical Rademacher complexity is defined as}

\begin{equation}
\widehat{\mathcal{R}}_{\mathcal{D}_{n}}(\mathcal{F})
=
\mathbb{E}_{\sigma}
\left[
\sup_{f \in \mathcal{F}}
\frac{1}{n}
\sum_{i=1}^{n}
\sigma_i f(x_i)
\right],
\end{equation}

\textit{where $\sigma_i$ are independent Rademacher variables taking values in $\{-1,+1\}$. This quantity measures the ability of a hypothesis family to fit random signs on the observed samples and is commonly used to characterize the capacity of a function class.}

\textbf{Proposition 2 (Capacity of the Semantically Admissible Predictor Family).}

\textit{Let $h_v^{\theta}$ denote the visual encoder, $h_s$ the frozen text encoder, and $r_{\theta}$ the regression head. The vision-only predictor used at inference is written as}

\begin{equation}
f_{\theta}(I)
=
r_{\theta}\!\left(h_v^{\theta}(I)\right).
\end{equation}

\textit{Let $\mathcal{F}_{v}$ denote the family of vision-only predictors realizable by the visual branch. Given a training set $\mathcal{T}=\{(I_i,S_i,y_i)\}_{i=1}^{n}$, we define the empirical cross-modal discrepancy associated with $f_{\theta}$ as}

\begin{equation}
\Omega_{\mathcal{T}}(f_{\theta})
=
\gamma \,
\mathcal{L}_{fal}
\left(
h_v^{\theta}(I),h_s(S)
\right)
+
\lambda \,
\mathcal{L}_{eal}
\left(
h_v^{\theta}(I),h_s(S)
\right),
\end{equation}

\textit{where the two losses are evaluated over $\mathcal{T}$. For a tolerance level $\epsilon \geq 0$, the semantically admissible predictor family is defined as}

\begin{equation}
\mathcal{F}_{\epsilon}
=
\left\{
f_{\theta} \in \mathcal{F}_{v}
:
\Omega_{\mathcal{T}}(f_{\theta})
\leq
\epsilon
\right\}.
\end{equation}

\textit{Since $\mathcal{F}_{\epsilon}$ is a subset of $\mathcal{F}_{v}$, its empirical Rademacher complexity satisfies}

\begin{equation}
\widehat{\mathcal{R}}_{\mathcal{D}_{n}}
\left(
\mathcal{F}_{\epsilon}
\right)
\leq
\widehat{\mathcal{R}}_{\mathcal{D}_{n}}
\left(
\mathcal{F}_{v}
\right).
\end{equation}

The proof is provided in~\ref{supp-proof2}.

\textbf{Implications of Proposition 1 and Proposition 2.}
Proposition 1 concerns the geometry and diversity of the learned representation, whereas Proposition 2 concerns the capacity of the predictor family selected by semantic-consistency constraints. Accordingly,
\begin{equation}
\widehat{\mathcal{R}}_{\mathcal{D}_{n}}
\left(
\mathcal{F}_{\epsilon}
\right)
\leq
\widehat{\mathcal{R}}_{\mathcal{D}_{n}}
\left(
\mathcal{F}_{v}
\right)
\end{equation}
does not imply a reduction in representational intrinsic dimension. The analysis in ~\ref{supp:id-geometry} shows that the effective spectra of both the visual and joint representations expand after alignment, which is compatible with a restricted admissible predictor family because the two quantities characterize representation geometry and function-class capacity, respectively. Proposition 2 provides a structural characterization of the admissible solution set rather than an empirical capacity estimate or a universal generalization guarantee.

\section{Method}
\label{sec4:method}

\subsection{Describe-To-Score}
\label{sec4.1}
\begin{figure}[t]
    \centering
    \includegraphics[width=1\linewidth]{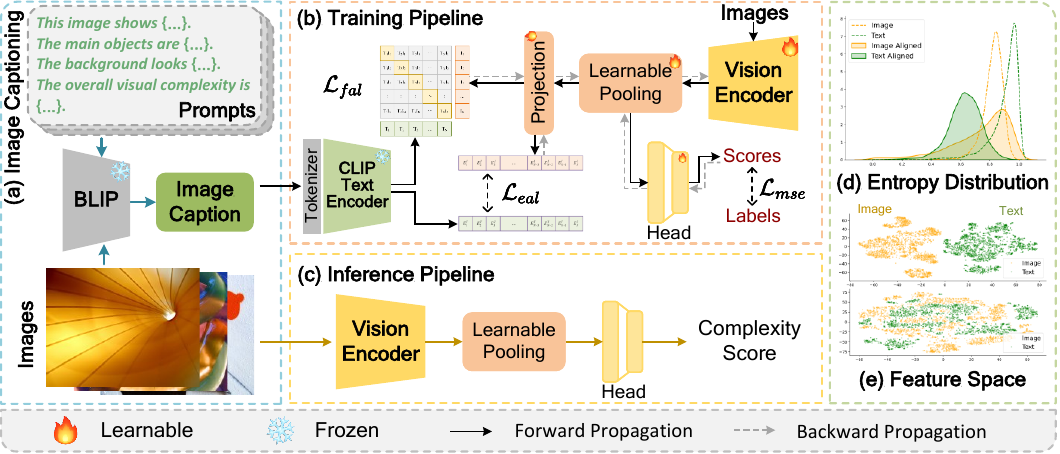}
    \caption{\textbf{Overall Architecture of the D2S framework.} (a) BLIP with a fill-in prompt generates captions. (b) Vision and text encoders extract features for regression loss $\mathcal{L}_{\mathrm{mse}}$ and the alignment losses $\mathcal{L}_{\mathrm{fal}}$, $\mathcal{L}_{\mathrm{eal}}$. (c) At inference, only image input is used for score prediction. (d) Entropy distribution before and after alignment (Dotted line: before alignment; solid line: after alignment).(e) Feature space before (top) and after (bottom) feature alignment.}
    \label{d2s}
\end{figure}

Our proposed D2S framework aims to jointly model low-level visual features and high-level semantic information through text guidance, thereby achieving robust ICA. Figure \ref{d2s} shows the overall workflow. Given an input image, a pre-trained BLIP first generates captions. The visual encoder then extracts visual features, while the text encoder extracts textual features. The vision-text alignment module of D2S consists of entropy distribution alignment and feature alignment. 

During training, textual information acts as a semantic teacher that reshapes the visual representation. However, \textbf{it is not directly used in the computation of the final complexity score}. The complexity score is produced solely from the aligned visual features through an MLP head. This design ensures that D2S benefits from semantic guidance during learning while remaining vision-only at inference.

\begin{wrapfigure}{r}{0.4\textwidth} 
\vspace{-10pt}
\includegraphics[width=1\linewidth]{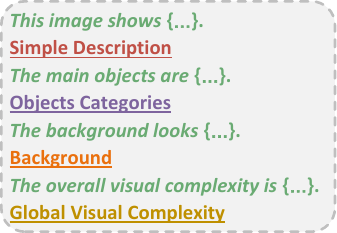}
\caption{Prompt template.}
\label{prompt}
\end{wrapfigure}

\textbf{Image captioning.} We adopt a well-designed fill-in-the-blank prompt template fed into BLIP-Large~\citep{li2022blip} to generate captions with images. We design four sentences (Figure \ref{prompt}) to obtain complementary descriptions focusing on \textbf{simple description}, \textbf{object categories}, \textbf{background}, and \textbf{global visual complexity}, respectively. For each image, the four prompts are applied sequentially, and their outputs are concatenated and cached before model optimization. The fourth prompt requests a qualitative description of global visual complexity without access to the ground-truth score or its numerical range. The cached captions are reused throughout training, and BLIP is not retained during D2S optimization. Representative generated captions and typical failure cases are shown in \ref{supp-capexp}.

\textbf{Vision encoder \& Text encoder.} The vision encoder in D2S is based on ResNet~\citep{he2016resnet}, but the original global pooling and fully connected layers are replaced with a learnable pooling layer that outputs a one-dimensional feature vector for each image. We employ the pre-trained CLIP~\citep{clip} text encoder as the text encoder in D2S, and use the CLIP tokenizer to segment each caption. The text encoder remains \textbf{frozen during training}, while gradients are updated only for the other components. Freezing the text encoder prevents semantic drift and maintains a stable semantic reference space.

\textbf{Vision-Text Connector.} Many prior works employ an MLP-style projection layer~\citep{chen2020simple} to map visual and textual features into the same dimensional space. Following this practice, we apply \textbf{one linear layer} as a projection for the vision encoder, aligning it with the dimensionality of the text features. This design ensures that text features can effectively guide visual features within a shared representation space. \textbf{Note, the projection will be discarded during inference.} Thus, the connector serves purely as a training-time alignment mechanism with no impact on model deployment.

\subsection{Entropy Distribution Alignment}
\label{sec4.2}
In the theoretical analysis, we pointed out that the weighted fusion of visual feature entropy $H_v^F(I)$ and semantic feature entropy $H_s^F(S)$ can better capture image complexity. However, during model learning, we observe that the empirical distributions of the two modalities show a noticeable bias (Figure \ref{d2s}(d), dotted line). Such bias may introduce additional uncertainty in the fusion stage and weaken the consistency of cross-modal complexity measurement. To address this issue, we propose entropy distribution alignment. By aligning the entropy distributions, D2S stabilizes the multimodal supervision signal and encourages a more coherent complexity-aware representation.

\textbf{Computing feature entropy.} 
To compute the entropy of visual and textual features, we follow the standard practice in information-theoretic analysis of neural representations. Given an intermediate feature vector $z \in \mathbb{R}^{d}$, we first normalize it using a temperature-scaled softmax to obtain a discrete probability distribution:
\begin{equation}
    p_{k} = \frac{\exp(z_{k} / \tau)}{\sum_{j=1}^{d} \exp(z_{j} / \tau)}, \qquad k = 1,\dots,d ,
\end{equation}
where $\tau$ is a temperature parameter that controls the smoothness of the distribution (\textit{we set $\tau=0.07$ unless otherwise specified}). The feature entropy is then computed using the Shannon entropy:
\begin{equation}
    H(z) = - \sum_{k=1}^{d} p_{k} \log p_{k}.
    \label{equation-entropy}
\end{equation}
For the visual and textual branches, we compute $H_{v}^{F}(I)$ and $H_{s}^{F}(S)$ by applying the above procedure to the outputs of the vision encoder and the text encoder, respectively. 
To avoid numerical instability, we add a small constant $\epsilon = 10^{-8}$ inside the logarithm. 
This entropy formulation treats the feature magnitude pattern as an empirical distribution over channels and provides a simple, consistent proxy for representational diversity.

\textbf{Entropy Buffer.} We establish and maintain two FIFO entropy buffers, $B_v$ and $B_s$, each with capacity $M$ to store visual and textual feature entropy.

\textbf{Buffer initialization.} Before training, we initialize the Momentum Model (MoM) as a parameter copy of D2S. The MoM receives no gradient updates and is updated exclusively through an exponential moving average of the online-model parameters.

\textbf{FIFO update.} At each iteration, we first use the MoM in inference mode to obtain the image and text features of all samples in the mini-batch, and then compute their entropy using Eq.(\ref{equation-entropy}). The visual and textual entropy are stored in their corresponding buffers. Once a buffer reaches half of its capacity, we begin to compute the entropy distribution alignment loss. We update the buffers by adding the new mini-batch samples and removing the oldest entries with the same number, thereby reducing distributional bias between old and new samples.

\textbf{Momentum refresh.} Furthermore, we update the MoM at each iteration using an exponential moving average (EMA) of the online-model parameters~\citep{he2020momentum}, and employ the updated MoM to refresh buffer entries with a \textit{refresh step}.

This mechanism keeps the entropy statistics temporally consistent and mitigates drift between modalities. The full algorithm of EAL is provided in \ref{supp-algorithm}.

\textbf{Entropy Distribution Alignment Loss $\mathcal{L}_{\mathrm{eal}}$.} We define the alignment loss using the energy distance~\citep{szekely2013energy}. Let
\begin{equation}
    V = \{v_{i}\}_{i=1}^{M}, \qquad S = \{s_{j}\}_{j=1}^{M}
\end{equation}
denote the entropy values stored in $B_v$ and $B_s$. The energy distance is defined as
\begin{equation}
    D_{E}^{2}(V, S) = 2\mathbb{E}[|V - S|] - \mathbb{E}[|V - V'|] - \mathbb{E}[|S - S'|],
\end{equation}
where $V,V'$ are i.i.d. samples from $B_v$ and $S,S'$ are i.i.d. samples from $B_s$. The empirical estimate is
\begin{equation}
    \mathcal{L}_{eal} = \frac{2}{M^2}\sum_{i=1}^{M}\sum_{j=1}^{M}|v_i-s_j| - \frac{1}{M(M-1)} \left( \sum_{i\ne i'} |v_i-v_{i'}| + \sum_{j\ne j'} |s_j-s_{j'}| \right).
\end{equation}

Minimizing this loss encourages the two modalities to share similar entropy statistics, improving their consistency during training.

\subsection{Feature Alignment}
\label{sec4.3}
To further exploit textual information, we align visual and textual features in a shared representation space. Inspired by CLIP~\citep{clip}, we introduce a contrastive loss that enforces correspondence between visual and textual representations. This encourages the visual encoder to absorb structural regularities present in the semantic space, complementing the entropy-level alignment of EAL. Let the outputs of the visual and text encoders be
\begin{equation}
    z_{v} = f_{v}(I), \qquad z_{s} = f_{s}(S),
    \label{eq-feat}
\end{equation}
which are then projected into the joint space through
\begin{equation}
    \tilde{z}_{v} = W_{v}z_{v}, \qquad \tilde{z}_{s} = W_{s}z_{s},
\end{equation}
where $W_{v}$ and $W_{s}$ denote projection parameters, and $W_{s}=\mathrm{I}$ in this work. With temperature parameter $\tau$, the InfoNCE loss~\citep{oord2019infonce} is:
\begin{equation}
    \mathcal{L}_{\mathrm{fal}} = -\frac{1}{N} 
    \sum_{i=1}^{N} 
    \log \frac{\exp(\mathrm{sim}(\tilde{z}_{v}^{i}, \tilde{z}_{s}^{i})/\tau)}
    {\sum_{j=1}^{N} \exp(\mathrm{sim}(\tilde{z}_{v}^{i}, \tilde{z}_{s}^{j})/\tau)}.
\end{equation}
This loss serves as a semantic regularizer that reshapes the visual representation manifold and suppresses task-irrelevant variations. The regression head is optimized using MSE $\mathcal{L}_{\mathrm{mse}}$. The final objective is
\begin{equation}
\mathcal{L} = \mathcal{L}_{\mathrm{mse}}
+ \lambda\, \mathcal{L}_{\mathrm{eal}}
+ \gamma\, \mathcal{L}_{\mathrm{fal}},
\end{equation}
where $\lambda$ and $\gamma$ control the relative weights of $\mathcal{L}_{\mathrm{eal}}$ and $\mathcal{L}_{\mathrm{fal}}$.

\section{Experiments}
\label{sec-exp-setting}
The reported results are averaged over three random seeds (42, 826, 1215), and we additionally report the standard deviation (std) in Table \ref{tab:1}. More training details are provided in Section \ref{supp-impl}.

\subsection{Setups}
\subsubsection{Implementation Details}
\label{supp-impl}
We implement our model in PyTorch. The ResNet backbone is initialized with ImageNet-pretrained~\citep{deng2009imagenet} weights from TIMM~\citep{rw2019timm}, and the BLIP caption generator is employed in its large configuration. The CLIP text encoder is also initialized with pretrained parameters. We train the model using the Adam~\citep{adam2014method} optimizer with weight decay 1e-3. The initial learning rate was set to 1e-3 and the batch size to 32. The cosine annealing~\citep{loshchilov2016sgdr} was used for scheduling, with a minimum learning rate of 2.5e-6. The temperature was set to 0.07. Training was performed on a single NVIDIA RTX 3090.

For the results in Table \ref{tab:2}, the training set was constructed by dividing the complexity score range $(0,1)$ into ten intervals and randomly sampling within each interval. Specifically, the SST@10 setting samples one image from each interval, whereas SST@50 samples five images from each interval. It is worth noting that, due to the scarcity of high-complexity images, the intended target of 50 samples per interval cannot be satisfied in the highest-complexity bin. In particular, the interval [0.9, 1.0] contains only 10 available training images, so the SST@500 protocol yields 460 realizable samples rather than 500.

For the results in Table \ref{tab:3}, we adopted a cross-dataset evaluation protocol, where the full datasets from the ICA benchmark, including Nagle4k~\citep{Nagle4k}, Savoias~\citep{saraee2020visual}, and VISC-C/I~\citep{visc}, were used as test sets.

For the results in Table \ref{tab:4}, we followed the common evaluation protocol in NR-IQA. After min-max normalizing the MOS scores of all images, each dataset was randomly partitioned into training, validation, and test subsets with a ratio of 6:2:2. We compared D2S with QPT~\citep{zhao2023QPT}, ARNIQA~\citep{agnolucci2024arniqa}, TOPIQ~\citep{chen2024topiq}, CDINet~\citep{zheng2024cdinet}, LoDa~\citep{xu2024loda}, ADTRS~\citep{alsaafin2024ADTRS}, VISGA~\citep{shi2025visga}, CoDI-IQA~\citep{liu2025codiiqa}, DGIQA~\citep{ramesh2025dgiqa} and RSFIQA~\citep{song2025RSFIQA}.

\begin{table}[t!]
\centering
\small
\caption{\textbf{Comparison with state-of-the-art image complexity assessment methods on IC9600.} Entropy buffer size is 2048 with a refresh step of 50 ($\sim$ 40 entropy were refreshed per iteration) and momentum 0.995. $\lambda$ and $\gamma$ are set to 5 and 0.01, respectively. $^\dagger$ denotes an unsupervised method. The results of ICNet and ICCORN are our re-implementation. The \textbf{bolded} portion marks the optimal outcome. The \underline{underlined} portion denotes the second-best. SRCC and PCC, larger $\uparrow$ is better. RMSE, RMAE, Params (M: million, B: billion), and Latency (ms: millisecond, s: second), smaller $\downarrow$ is better.}
\label{tab:1}
\begin{tabular}{l|cccccc}
\toprule
\multicolumn{1}{l|}{\textbf{Method}} & \textbf{SRCC} & \textbf{PCC} & \textbf{RMSE}  & \textbf{RMAE} & \textbf{Params} & \textbf{Latency} \\ \midrule
$\text{MoCo}^\dagger$~\citep{he2020momentum} & 0.759 & 0.748 & - & - & - &- \\
$\text{SAE}^\dagger$~\citep{saraee2020visual}  & 0.865 & 0.860 & 0.074 & 0.240 & - & - \\
$\text{CLIC}^\dagger$~\citep{liu2025clic} & 0.866 & 0.858 & - & - & - & - \\
$\text{CLICv2}^\dagger$~\citep{liu2025clicv2} & 0.879 & 0.870 & - & - & - & - \\ \midrule
HyperIQA~\citep{su2020hyperiqa} & 0.935 & 0.935 & 0.067 & 0.229 & 27.38M & 29.638ms \\
P2P-FM~\citep{ying2020p2pfm}   & 0.940 & 0.936 & 0.056 & 0.208 & - & - \\
TOPIQ~\citep{chen2024topiq}   & 0.938 & 0.944 & 0.049 & - & 45.20M & 14.434ms \\
ICNet~\citep{feng2022ic9600} & 0.9446 & 0.9470 & 0.0582 & 0.2156 & \underline{20.33M} & \underline{9.369ms} \\
std & ±.0029 & ±.0011 & ±.0021 & ±.0051 &  &    \\
ICCORN~\citep{guo2023iccorn} & 0.9455 & 0.9490 & 0.0526 & 0.2085 & 22.31M & 13.973ms \\
std & ±.0011 & ±.0018 & ±.0013 & ±.0053 &  &  \\
MICM~\citep{li2025micm} & 0.943 & 0.953 & 0.060 & - & $\sim$11B & $\sim$180s \\ \midrule
D2S-R18 (ours) & \underline{0.9509} & \underline{0.9544} & \textbf{0.0495} & \textbf{0.1962} & \textbf{13.02M} & \textbf{4.573ms} \\
std & ±.0026 & ±.0009 & ±.0016 & ±.0050 &  &  \\
D2S-R50 (ours)  & \textbf{0.9541} & \textbf{0.9576} & \underline{0.0496} & \underline{0.1963} & 38.72M & 10.738ms \\
std & ±.0004 & ±.0010 & ±.0028 & ±.0062 &  & \multicolumn{1}{l}{} \\
\bottomrule
\end{tabular}
\end{table}

\subsubsection{Datasets}
\label{supp-data}
The datasets used in our study are drawn from two tasks: image complexity assessment (ICA) and image quality assessment (IQA). The ICA datasets include IC9600, Savoias, PASCAL VOC\_4000 (Nagle4k), and VISC-C/I. Among them, IC9600 is employed to validate the effectiveness of our method, while Savoias, Nagle4k, and VISC-C/I are used to evaluate cross-dataset generalization. The IQA datasets, consisting of KADID-10K, KonIQ-10K, and TID2013, are utilized to assess the cross-task transferability of the proposed approach.

\subsubsection{Evaluation Metrics}
\label{supp-metrics}
We adopt Spearman’s Rank Correlation Coefficient (SRCC), Pearson’s Linear Correlation Coefficient (PCC), Root Mean Square Error (RMSE), and Relative Mean Absolute Error (RMAE) to evaluate the predictive performance of our model. In addition, we report the number of parameters (Params) and inference latency (Latency) to assess computational efficiency.

\subsection{Main Results}
\label{sec-exp-main}

\begin{figure}[t]
\centering
\begin{minipage}[t]{0.48\textwidth}
\centering
\includegraphics[width=1\linewidth]{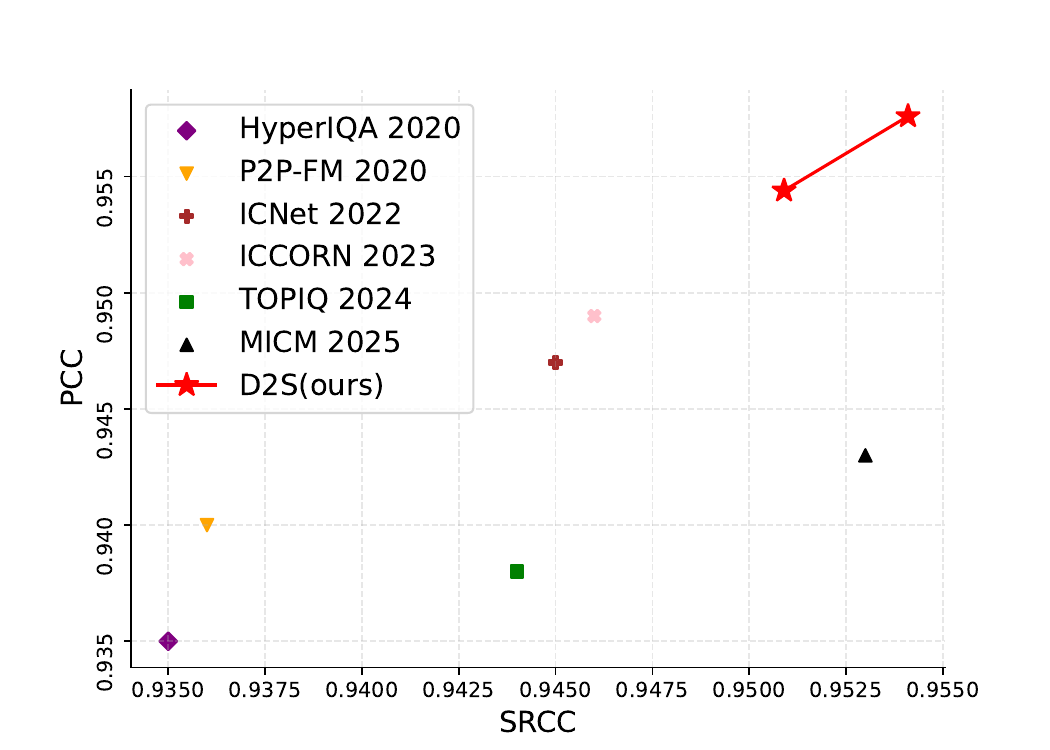}
\caption{Accuracy comparison on IC9600 using SRCC and PCC. Points closer to the upper-right corner indicate stronger correlation performance.}
\label{PCC_SRCC}
\end{minipage}\hfill
\begin{minipage}[t]{0.48\textwidth}
\centering
\includegraphics[width=1\linewidth]{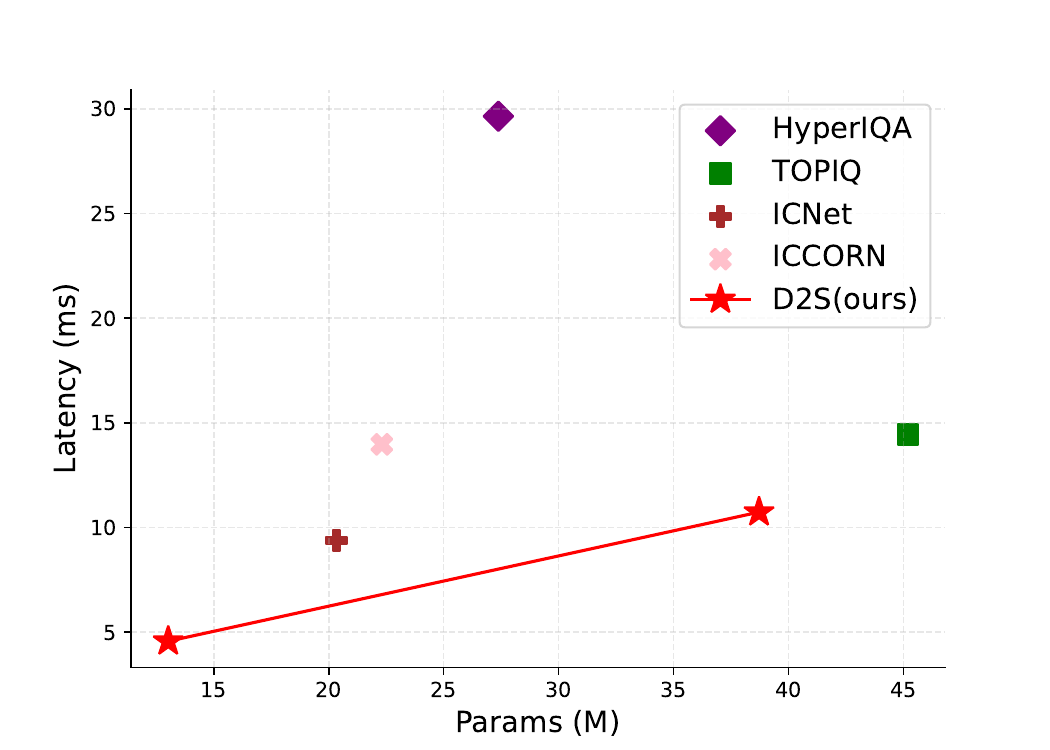}
\caption{Efficiency comparison on IC9600 in terms of model parameters and inference latency. Methods closer to the lower-left corner are more efficient.}
\label{Latency_Params}
\end{minipage}
\end{figure}

\textbf{Comparison with State-of-the-Art Methods.} We benchmark D2S against both unsupervised and supervised ICA methods on IC9600, with results summarized in Table \ref{tab:1}. Unsupervised methods show limited performance, whereas supervised models benefit substantially from labeled training data. Among the strongest baselines, MICM achieves competitive correlation but requires extremely large models and high inference cost, limiting its practical utility.

\begin{table}[t]
\centering
\small
\setlength{\tabcolsep}{14pt}
\caption{\textbf{Paired bootstrap significance test on IC9600.} Mean $\Delta$ denotes the performance difference between D2S-R50 and the reproduced baseline. Positive $\Delta$ indicates improvement for SRCC and PCC, while negative $\Delta$ indicates improvement for RMSE and RMAE. The 95\% confidence intervals (95\% CI) and one-sided empirical $p$-values are computed using 5,000 paired bootstrap resamplings. When no resampled trial falls on the non-improvement side, the $p$-value is reported as $<0.0002$.}
\label{tab:bootstrap_sig}
\begin{tabular}{llccc}
\toprule
\textbf{Baseline} & \textbf{Metric} & \textbf{Mean $\Delta$} & \textbf{95\% CI} & \textbf{$p$-value} \\
\midrule
ICNet  & SRCC & +0.0094 & [0.0054, 0.0136] & $<0.0002$ \\
ICNet  & PCC  & +0.0088 & [0.0058, 0.0119] & $<0.0002$ \\
ICNet  & RMSE & -0.0032 & [-0.0046, -0.0019] & $<0.0002$ \\
ICNet  & RMAE & -0.0071 & [-0.0097, -0.0044] & $<0.0002$ \\
\midrule
ICCORN & SRCC & +0.0086 & [0.0063, 0.0109] & $<0.0002$ \\
ICCORN & PCC  & +0.0086 & [0.0046, 0.0126] & $<0.0002$ \\
ICCORN & RMSE & -0.0030 & [-0.0090, 0.0031] & $0.3310$ \\
ICCORN & RMAE & -0.0122 & [-0.0282, 0.0038] & $0.1347$ \\
\bottomrule
\end{tabular}
\end{table}

\begin{table}[t]
\centering
\scriptsize
\caption{\textbf{Performance of small samples training on IC9600.} SST@X denotes few-shot training with X samples. Epoch: 5. Hyper-parameters are same as Table \ref{tab:1}. SRCC and PCC, larger $\uparrow$ is better. RMSE and RMAE, smaller $\downarrow$ is better.}
\label{tab:2}
\begin{tabular}{c|cccc|cccc}
\toprule
\textbf{Method} & \textbf{SRCC}  & \textbf{PCC}  & \textbf{RMSE}  & \textbf{RMAE}  & \textbf{SRCC}  & \textbf{PCC} & \textbf{RMSE}  & \textbf{RMAE}  \\ \midrule
\textbf{} & \multicolumn{4}{c|}{SST@10} & \multicolumn{4}{c}{SST@50} \\ \cmidrule{2-9} 
ICNet & 0.5178 & 0.5057 & 0.1527 & 0.3429 & 0.5644 & 0.5864 & 0.1514 & 0.3422 \\ 
ICCORN & 0.5221 & 0.5102 & 0.1497 & 0.3393 & 0.5704 & 0.5914 & 0.1484 & 0.3386 \\
D2S-R18 & 0.6664 & 0.6839 & 0.1302 & 0.3199 & 0.8174 & 0.8332 & \textbf{0.1030} & \textbf{0.2858} \\
D2S-R50 & \textbf{0.6864} & \textbf{0.6972} & \textbf{0.1190} & \textbf{0.3055} & \textbf{0.8515} & \textbf{0.8528} & 0.1133 & 0.3025 \\ \cmidrule{2-9} 
\textbf{} & \multicolumn{4}{c|}{SST@100} & \multicolumn{4}{c}{SST@500} \\ \cmidrule{2-9} 
ICNet & 0.8239 & 0.8259 & 0.1668 & 0.3820 & 0.8942 & 0.9054 & 0.0830 & 0.2582 \\ 
ICCORN & 0.8276 & 0.8294 & 0.1605 & 0.3747 & 0.8977 & 0.9080 & 0.0791 & 0.2527 \\
D2S-R18 & 0.8552 & 0.8547 & \textbf{0.0964} & \textbf{0.2787} & 0.9115 & 0.9182 & \textbf{0.0707} & \textbf{0.2326} \\
D2S-R50 & \textbf{0.8680} & \textbf{0.8731} & 0.1092 & 0.2913 & \textbf{0.9182} & \textbf{0.9246} & 0.0709 & 0.2349 \\ \bottomrule
\end{tabular}
\end{table}

By contrast, D2S achieves a more favorable accuracy-efficiency trade-off. As shown in Figure \ref{PCC_SRCC}, both D2S variants lie in the upper-right region of the SRCC--PCC plane, indicating strong correlation performance; D2S-R50 attains the best overall results, while D2S-R18 remains highly competitive with a much smaller model size. The efficiency comparison in Figure \ref{Latency_Params} further shows that D2S occupies a favorable low-latency, moderate-parameter region relative to prior methods. In particular, D2S-R18 offers strong accuracy with clearly reduced inference cost, whereas D2S-R50 achieves the highest SRCC and PCC among the compared supervised ICA models while retaining practical inference efficiency.

\textbf{Statistical reliability.} To further assess statistical reliability, we performed paired bootstrap testing with 5,000 resamplings on the IC9600 test set. As shown in Table~\ref{tab:bootstrap_sig}, D2S-R50 significantly improves SRCC and PCC over both reproduced baselines, including the strongest visual baseline ICCORN. The error-based metrics also decrease in mean value, although the reductions over ICCORN are not statistically significant. Together with the mean and standard deviation over three random seeds in Table~1, these results suggest that the main correlation gains of D2S are unlikely to be caused by training randomness or test-sample fluctuation.

\begin{figure}[t]
\centering
\begin{minipage}[t]{0.49\textwidth}
\centering
\includegraphics[width=1\linewidth]{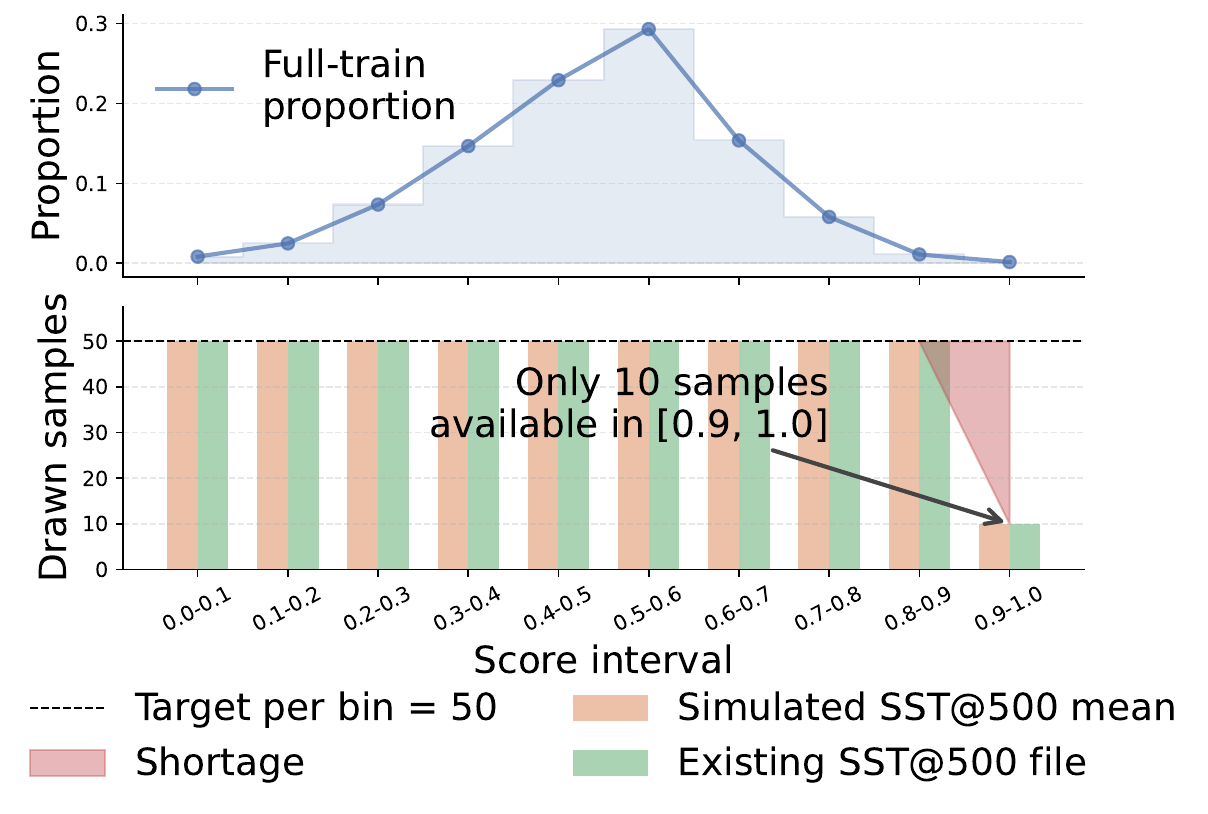}
\caption{\textbf{Interval occupancy in the SST@500 protocol.} SST@500 yields only 460 realizable samples, as $[0.9, 1.0]$ contains just 10 training images.}
\label{sst500_bias_summary}
\end{minipage}\hfill
\begin{minipage}[t]{0.49\textwidth}
\centering
\includegraphics[width=1\linewidth]{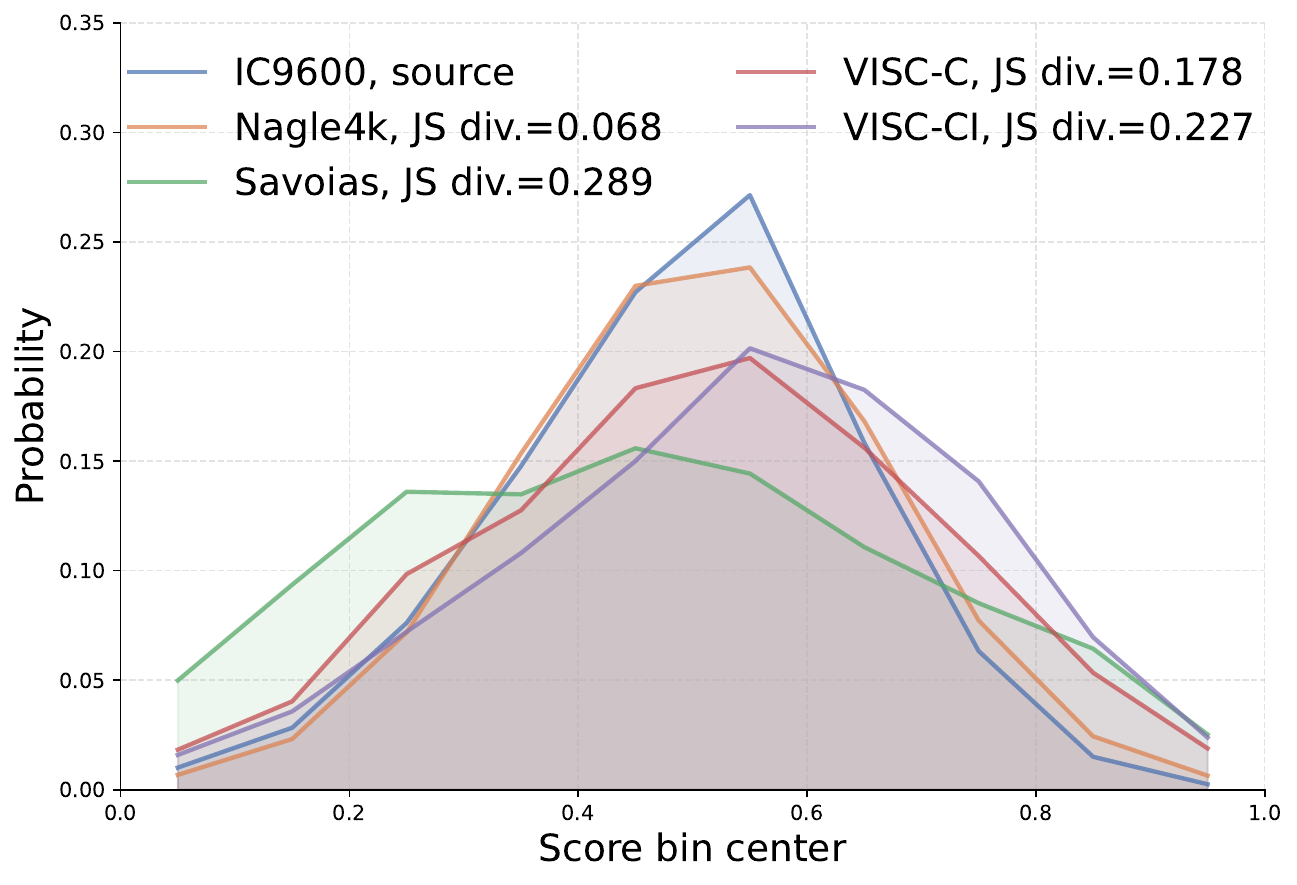}
\caption{\textbf{Score-distribution shift between IC9600 and target datasets.} JS div. from IC9600 is shown in the legend; larger values indicate greater mismatch.}
\label{cross_dataset_shift_hist}
\end{minipage}
\end{figure}

\textbf{Small Samples Training (SST).} We evaluate the few-shot behavior of D2S by randomly sampling 10, 50, 100, and 500 training images from IC9600 (Table \ref{tab:2}). Note that SST@500 is not a uniformly populated interval-sampling setting. As shown in Figure \ref{sst500_bias_summary}, it yields only 460 realizable samples rather than 500 because the highest-complexity interval $[0.9,1.0]$ contains only 10 training images. Thus, SST@500 should be interpreted as a long-tailed sampling regime biased toward low- and mid-complexity regions.

Despite this limitation, Table \ref{tab:2} still reveals clear relative trends under low-data supervision. Performance improves steadily with sample size, especially in SRCC and PCC. With only 10 samples, D2S-R50 reaches 0.6864 SRCC, substantially outperforming ICNet (0.5178). With 50--100 samples, performance increases rapidly (e.g., 0.8680 at SST@100). Under SST@500, both D2S variants exceed 0.91 in SRCC and PCC while reducing RMSE and RMAE. In contrast, ICNet and ICCORN deteriorate more noticeably in limited-data settings. These results show that D2S remains effective in low-data regimes, although the SST@500 results should be interpreted together with the long-tailed interval occupancy described above.

\begin{table}[t]
\centering
\scriptsize
\caption{\textbf{Cross-dataset generalization of D2S on ICA datasets.} Epoch: 5. Hyper-parameters are same as Table \ref{tab:1}. SRCC and PCC, larger $\uparrow$ is better. RMSE and RMAE, smaller $\downarrow$ is better.}
\label{tab:3}
\begin{tabular}{c|cccc|cccc}
\toprule
\textbf{Method} & \textbf{SRCC} & \textbf{PCC} & \textbf{RMSE} & \textbf{RMAE} & \textbf{SRCC} & \textbf{PCC} & \textbf{RMSE} & \textbf{RMAE} \\ \midrule
\textbf{} & \multicolumn{4}{c|}{Nagle4k~\citep{Nagle4k}} & \multicolumn{4}{c}{VISC-C~\citep{visc}} \\ \cmidrule{2-9} 
ICNet & 0.7851 & 0.7666 & 0.1082 & 0.2917 & 0.7219 & 0.7111 & 0.1415 & 0.3367 \\ 
ICCORN & 0.7905 & 0.7706 & \textbf{0.1070} & \textbf{0.2891} & 0.7251 & 0.7155 & 0.1407 & 0.3340 \\
D2S-R18 & 0.7976 & 0.7748 & 0.1102 & 0.2949 & 0.7291 & 0.7165 & \textbf{0.1399} & \textbf{0.3350} \\
D2S-R50 & \textbf{0.7976} & \textbf{0.7765} & 0.1126 & 0.2989 & \textbf{0.7317} & \textbf{0.7170} & 0.1421 & 0.3375 \\ \cmidrule{2-9} 
\textbf{} & \multicolumn{4}{c|}{Savoias~\citep{saraee2020visual}} & \multicolumn{4}{c}{VISC-CI~\citep{visc}} \\ \cmidrule{2-9} 
ICNet & 0.6813 & 0.6793 & 0.1721 & 0.3736 & 0.6802 & 0.6876 & \textbf{0.1549} & \textbf{0.3571} \\ 
ICCORN & 0.6835 & 0.6811 & 0.1719 & 0.3720 & 0.6825 & 0.6924 & 0.1563 & 0.3591 \\
D2S-R18 & 0.6780 & 0.6825 & 0.1706 & 0.3717 & \textbf{0.6853} & \textbf{0.6982} & 0.1580 & 0.3617 \\
D2S-R50 & \textbf{0.6845} & \textbf{0.6882} & \textbf{0.1700} & \textbf{0.3713} & 0.6828 & 0.6963 & 0.1622 & 0.3660 \\ \bottomrule
\end{tabular}
\end{table}

\textbf{Cross-Dataset Generalization.} To evaluate generalization, we train D2S on IC9600 and directly test on Savoias, Nagle4k, and VISC-C/I without fine-tuning (Table \ref{tab:3}). Across most metrics, both D2S variants remain competitive with or superior to ICNet and ICCORN. The gains are clearer on Nagle4k and VISC-C, whereas all methods remain below 0.70 on Savoias and VISC-C/I. To better understand this behavior, we compare the score distributions of the target datasets against IC9600 in Figure \ref{cross_dataset_shift_hist}, where the Jensen-Shannon divergence (JS div.) is reported in the legend. Nagle4k shows the smallest shift, while Savoias and VISC-CI exhibit substantially larger mismatch. This trend is consistent with the transfer results in Table \ref{tab:3} and suggests that cross-dataset performance is constrained by annotation and distribution differences in addition to representation quality. Overall, text-guided alignment improves robustness, but cross-dataset transfer remains limited under substantial domain and label shifts. We further provide full-data training results on these ICA benchmarks in \ref{supp-full data}, where D2S remains competitive when trained directly on each target dataset.

\begin{table}[t]
\centering
\footnotesize
\caption{\textbf{Cross-task transfer of D2S to NR-IQA.} D2S were trained for 5 epochs with input resolution 384. Method references and implementation details are provided in Section \ref{supp-impl}.}
\label{tab:4}
\begin{tabular}{@{}cc|cccccc@{}}
\toprule
\multirow{2}{*}{\textbf{Method}} & \multirow{2}{*}{\textbf{Source}} & \multicolumn{2}{c}{\textbf{KADID-10K}} & \multicolumn{2}{c}{\textbf{KonIQ-10K}} & \multicolumn{2}{c}{\textbf{TID2013}} \\ \cmidrule(l){3-8} 
 &  & \textbf{SRCC} $\uparrow$ & \textbf{PCC} $\uparrow$ & \textbf{SRCC} $\uparrow$ & \textbf{PCC} $\uparrow$ & \textbf{SRCC} $\uparrow$ & \textbf{PCC} $\uparrow$ \\ \midrule
QPT & CVPR'23 & 0.925 & 0.928 & - & - & 0.895 & 0.914 \\
ARNIQA & WACV'24 & 0.908 & 0.912 & - & - & 0.880 & 0.901 \\
TOPIQ & TIP'24 & 0.921 & 0.924 & 0.574 & 0.657 & 0.870 & 0.884 \\
CDINet & TMM'24 & 0.920 & 0.919 & 0.865 & 0.880 & 0.898 & 0.908 \\
LoDa & CVPR'24 & 0.876 & 0.899 & 0.932 & 0.944 & 0.869 & 0.901 \\
ADTRS & ICIP'24 & - & - & 0.905 & 0.918 & 0.878 & 0.897 \\
VISGA & TCSVT'25 & 0.919 & 0.925 & 0.930 & 0.937 & 0.901 & 0.914 \\
CoDI-IQA & ArXiv'25 & 0.936 & 0.940 & 0.902 & 0.917 & 0.901 & 0.916 \\
DGIQA & ArXiv'25 & 0.943 & 0.945 & \textbf{0.934} & \textbf{0.942} & 0.934 & 0.940 \\
RSFIQA & ArXiv'25 & \underline{0.953} & \underline{0.954} & \underline{0.934} & \underline{0.940} & \textbf{0.951} & \textbf{0.959} \\ \midrule
D2S-R18& \multicolumn{1}{c|}{\multirow{2}{*}{ours}} & 0.952 & 0.953 & 0.901 & 0.922 & \underline{0.941} & \underline{0.938} \\
D2S-R50& \multicolumn{1}{c|}{} & \textbf{0.958} & \textbf{0.959} & 0.900 & 0.925 & 0.938 & 0.935 \\ \bottomrule
\end{tabular}
\end{table}

\textbf{Cross-Task Transfer.} We further assess transferability by applying D2S to NR-IQA (Table \ref{tab:4}). Training and dataset details are provided in \ref{supp-impl} and \ref{supp-data}. D2S achieves competitive performance relative to recent NR-IQA methods, and is especially strong on KADID-10K. On KADID-10K, D2S-R50 obtains 0.958 SRCC and 0.959 PCC, the best among all compared methods. On KonIQ-10K, DGIQA remains the strongest, but D2S achieves competitive values near 0.90 SRCC. On TID2013, both variants exceed 0.938. These results suggest that semantic alignment benefits not only ICA but also related perceptual prediction tasks that depend on visual–semantic coherence.

\subsection{Ablations}
\textbf{Main components.} We evaluate the effect of AttnPool, EAL, and FAL by incrementally adding each module (Table \ref{tab:5}). AttnPool improves performance by enabling fine-grained feature aggregation. FAL provides moderate gains in correlation metrics, while EAL mainly improves error reduction and stability by controlling cross-modal entropy bias. When combined, D2S achieves the best performance (SRCC 0.9499, PCC 0.9540) with lower RMSE and RMAE. The complementary behavior of these modules highlights that semantic alignment must occur both at the representation level (FAL) and the distributional level (EAL) to maximize performance.

\begin{table}[t]
\centering
\small
\caption{\textbf{Ablation study of the main components in D2S on IC9600.} Each case (a $\sim$ e) corresponds to different combinations of these modules.}
\label{tab:5}
\begin{tabular}{c|ccc|cccc}
\toprule
\textbf{Case} & \textbf{AttnPool} & \textbf{EAL} & \textbf{FAL} & \textbf{SRCC} $\uparrow$ & \textbf{PCC} $\uparrow$ & \textbf{RMSE} $\downarrow$ & \textbf{RMAE} $\downarrow$ \\ \midrule
(a) & × & × & × & 0.9396 & 0.9430 & 0.0547 & 0.2052 \\
(b) & \checkmark & × & × & 0.9446 & 0.9476 & 0.0541 & 0.2050 \\
(c) & \checkmark & \checkmark & × & 0.9467 & 0.9503 & 0.0546 & 0.2070 \\
(d) & \checkmark & × & \checkmark & 0.9473 & 0.9510 & 0.0551 & 0.2086 \\
(e) & \checkmark & \checkmark & \checkmark & \textbf{0.9499} & \textbf{0.9540} & \textbf{0.0472} & \textbf{0.1906} \\ \bottomrule
\end{tabular}
\end{table}

\begin{table}[t]
\centering
\small
\setlength{\tabcolsep}{3pt}
\caption{\textbf{Ablation study of EAL hyperparameters and training-resource overhead on IC9600.} Training time and peak memory are measured on a single NVIDIA RTX A2000 under identical batch-size and precision settings. $M$, mom., and steps denote the buffer capacity, momentum coefficient, and refresh interval, respectively.}
\label{tab:6}
\begin{tabular}{ccc|cccccc}
\toprule
\textbf{$M$} & \textbf{mom.} & \textbf{steps} & \textbf{SRCC} $\uparrow$ & \textbf{PCC} $\uparrow$ & \textbf{RMSE} $\downarrow$ & \textbf{RMAE} $\downarrow$ & \textbf{Time (h)} $\downarrow$ & {\color{orange}Mem. (GB) $\downarrow$} \\ \midrule
\multicolumn{3}{c}{Vision-only} & 0.9468 & 0.9495 & 0.0522 & 0.2010 & 0.6 &4.428 \\ \midrule
\multirow{4}{*}{2048} & 0.99 & \multirow{4}{*}{50} & 0.9487 & 0.9533 & 0.0497 & 0.1955 & - & - \\
 & 0.995 &  & \textbf{0.9508} & \textbf{0.9545} & 0.0496 & 0.1962 & - & - \\
 & 0.999 &  & 0.9499 & 0.9540 & \textbf{0.0472} & \textbf{0.1906}  & - & - \\
 & 0.9999 &  & 0.9484 & 0.9527 & 0.0497 & 0.1960  & - & - \\ \midrule
512 & \multirow{4}{*}{0.995} & \multirow{4}{*}{50} & 0.9494 & 0.9531 & 0.0519 & 0.2016 & \textbf{0.8} & 5.528 \\
1024 &  &  & 0.9493 & 0.9533 & 0.0516 & 0.2008 & \underline{0.9} & 5.528 \\
2048 &  &  & \textbf{0.9508} & \textbf{0.9545} & \textbf{0.0496} & \textbf{0.1962} & 1.1 & 5.529 \\
4096 &  &  & 0.9501 & 0.9530 & 0.0529 & 0.2041 & 1.9 & 5.530 \\ \midrule
\multirow{3}{*}{2048} & \multirow{3}{*}{0.995} & 16 & 0.9499 & 0.9539 & 0.0513 & 0.2008 & - & -  \\
 &  & 50 & \textbf{0.9508} & \textbf{0.9545} & \textbf{0.0496} & \textbf{0.1962} & - & -  \\
 &  & 128 & 0.9496 & 0.9539 & 0.0502 & 0.1973 & - & -  \\ \bottomrule
\end{tabular}
\end{table}

\begin{figure}[t]
\centering
\begin{minipage}[t]{0.49\textwidth}
\centering
\includegraphics[width=1\linewidth]{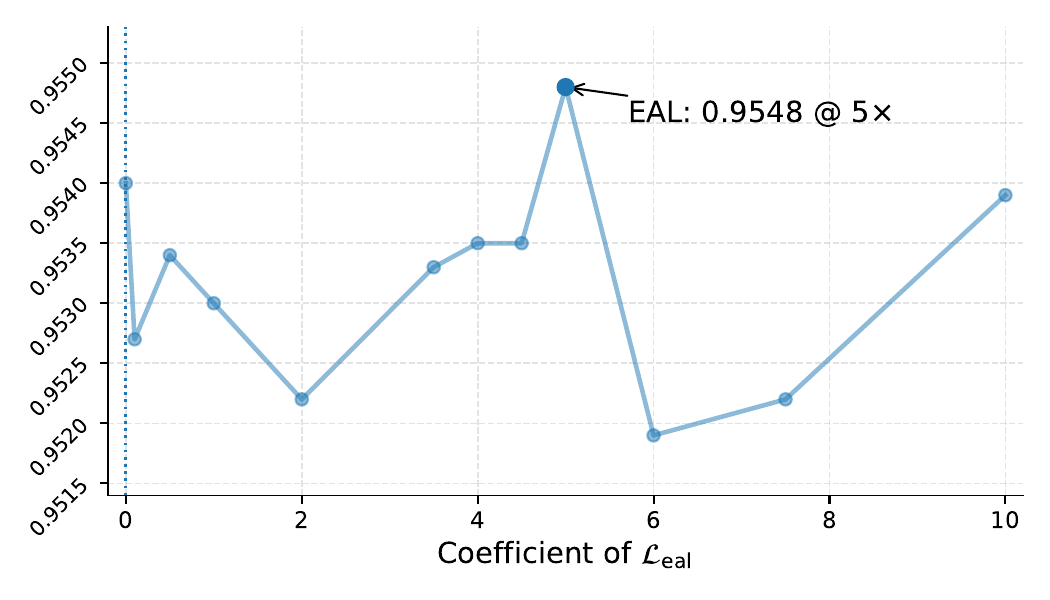}
\caption{\textbf{EAL coefficient ablation.} D2S is trained for 20 epochs with ResNet18 as the backbone, while fixing the FAL coefficient at 0.01. The best result is achieved when the EAL coefficient equals 5.}
\label{plcc_vs_eal_coeff}
\end{minipage}\hfill
\begin{minipage}[t]{0.49\textwidth}
\centering
\includegraphics[width=1\linewidth]{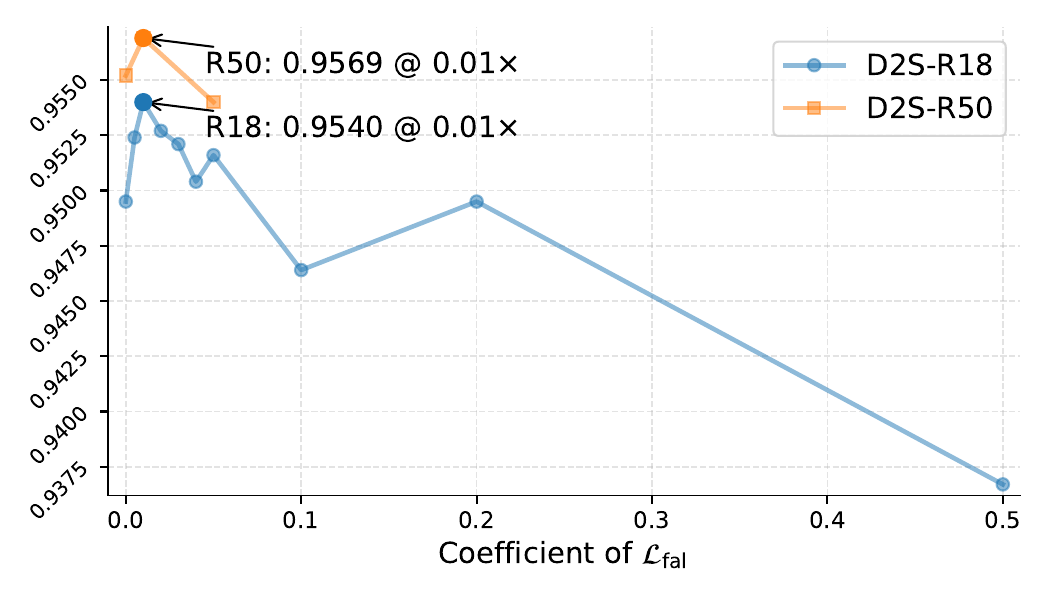}
\caption{\textbf{FAL coefficient ablation.} D2S is trained for 20 epochs with ResNet18 and ResNet50 as backbones, while fixing the EAL coefficient at 5. The best FAL coefficient is 0.01.}
\label{plcc_vs_fal_coeff}
\end{minipage}
\end{figure}

\textbf{EAL and FAL coefficient.} We fix ResNet18 as the backbone and vary the EAL coefficient while keeping FAL coefficient at 0.01, buffer size 2048, momentum 0.995, and refresh step 50. As shown in Figure \ref{plcc_vs_eal_coeff}, the best PCC (0.9548) occurs at EAL coefficient 5. Table \ref{tab:6} further shows the robustness of D2S to EAL hyper-parameters. The performance variation remains small across a wide range of settings, indicating that EAL benefits primarily from relative rather than absolute entropy alignment strength.

We vary the FAL coefficient while fixing EAL at its optimal configuration. Experiments using ResNet18 and ResNet50 show that the best performance is obtained at a FAL coefficient of 0.01 (Figure \ref{plcc_vs_fal_coeff}). Larger coefficients occasionally destabilize training due to excessive semantic pulling, while smaller values reduce the regularization effect.

Under the default setting with $M=2048$, training time increases from 0.6 to 1.1 h and peak memory from 4.428 to 5.529 GB. Peak memory remains nearly unchanged as $M$ increases from 512 to 4096, consistent with the scalar storage used by the entropy buffers.

\textbf{Prompt Combination Matrix.} Table~\ref{tab:prompt-matrix} evaluates individual prompts and representative multi-prompt combinations among the four prompts in Figure \ref{prompt}. The explicit global-complexity prompt $P_4$ alone yields an SRCC of 0.9456, below the no-prompt result of 0.9468, although it produces lower RMSE and RMAE. Adding $P_4$ to $[P_1,P_2,P_3]$ results in limited changes in SRCC and PCC, while the complete prompt set achieves the best PCC, RMSE, and RMAE and remains close to the best SRCC. The performance is therefore supported by complementary semantic descriptions rather than dominated by the global-complexity prompt.

\begin{table}[t]
\centering
\small
\caption{\textbf{Ablation of different prompt combinations.} All experiments are conducted on D2S-R18. Pi denotes the i-th prompt in Figure \ref{prompt}. $[:, :]$ indicates the concatenation or combination of multiple prompts. \textbf{Bold} indicates the best overall performance. \underline{Underline} indicates the best performance achieved by any single prompt.}
\label{tab:prompt-matrix}
\begin{tabular}{l|cccc}
\toprule
\textbf{Prompts} & \textbf{SRCC} $\uparrow$ & \textbf{PCC} $\uparrow$ & \textbf{RMSE} $\downarrow$ & \textbf{RMAE} $\downarrow$ \\
\midrule
None    & 0.9468 & 0.9495 & 0.0522 & 0.2010 \\
P1      & \underline{0.9511} & \underline{0.9533} & 0.0547 & 0.2082 \\
P2      & 0.9458 & 0.9505 & 0.0539 & 0.2053 \\
P3      & 0.9486 & 0.9520 & 0.0536 & 0.2053 \\
P4      & 0.9456 & 0.9507 & \underline{0.0509} & \underline{0.1986} \\
\midrule
$[$P1, P2, P3$]$    & 0.9487 & 0.9526 & 0.0521 & 0.2017 \\
$[$P1, P2, P4$]$    & 0.9489 & 0.9527 & 0.0523 & 0.2019 \\
$[$P1, P3, P4$]$    & 0.9485 & 0.9522 & 0.0517 & 0.2006 \\
$[$P2, P3, P4$]$    & 0.9489 & 0.9532 & 0.0522 & 0.2018 \\
$[$P1, P2, P3, P4$]$ & \textbf{0.9509} & \textbf{0.9544} & \textbf{0.0495} & \textbf{0.1962} \\
\bottomrule
\end{tabular}
\end{table}

\textbf{Impact of Different Caption Generators.} We further evaluate full D2S training with captions from Florence-2 and Qwen3-VL-4B (Table \ref{tab9}). On IC9600, stronger captioners do not significantly outperform BLIP. This suggests that D2S is not tied to a specific captioner and that approximate semantics are sufficient to act as a soft regularizer for guiding the visual encoder. A consolidated modality-level comparison among text-only, vision-only, and vision-text methods is provided in \ref{supp-discuss}.

\begin{table}[t]
\centering
\footnotesize
\caption{\textbf{Comparison of different caption generators.} These experiments are based on D2S-R18. BLIP and Qwen generate captions using the prompts shown in Figure \ref{prompt}.}
\begin{tabular}{ll|cccc}
\toprule
\textbf{Captioner} & \textbf{Prompts} & \textbf{SRCC} $\uparrow$ & \textbf{PCC} $\uparrow$ & \textbf{RMSE} $\downarrow$ & \textbf{RMAE} $\downarrow$ \\
\midrule
Florence2 & caption               & 0.9456     & 0.9501     & 0.0535     & 0.2044     \\
Florence2 & detailed caption      & 0.9469     & 0.9506     & 0.0534     & 0.2044     \\
Florence2 & more detailed caption & 0.9441     & 0.9484     & 0.0524     & 0.2015     \\
BLIP Large & Figure \ref{prompt} prompts        & \underline{0.9509}     & \underline{0.9544}     & \textbf{0.0495} & \textbf{0.1962} \\
Qwen3-VL-4B & Figure \ref{prompt} prompts       & \textbf{0.9519} & \textbf{0.9552} & \underline{0.0503}     & \underline{0.1976}     \\
\bottomrule
\label{tab9}
\end{tabular}
\end{table}

\textbf{Effect of Incorrect or Corrupted Captions.} To quantify the robustness of D2S against degraded semantic supervision, we evaluate four forms of corrupted captions: full shuffling, partial shuffling, random sentences, and fixed meaningless words. Table~\ref{tab:badcap} shows that mild corruption produces only small drops, while removing semantics entirely (case d) leads to a clear decline, confirming that semantic structure (rather than mere text tokens) drives the improvement in D2S.

\textbf{Performance Using Only Captions.} To disentangle the contribution of textual information, we train models using captions alone without image input. As shown in Table \ref{tab:7} (left), performance ranges from 0.7095 (CapI) to 0.8260 (BLIP) in SRCC. This confirms that captions contain meaningful complexity cues but cannot fully capture visual diversity, reinforcing the complementary nature of multimodal fusion.

\begin{table}[t]
\centering
\small
\caption{\textbf{Effect of incorrect captions on IC9600 performance.} ‘--’ denotes the original set of four prompts used in Figure \ref{prompt}. ‘m=1.0’ indicates that all words are fully shuffled, and ‘m=0.5’ indicates that half of the words are shuffled. ‘rs’ refers to grammatically valid random sentences. ‘mw’ refers to meaningless token sequences such as ‘any any any …’}
\label{tab:badcap}
\begin{tabular}{c|c|cccc}
\toprule
\textbf{Case} & \textbf{Method} & \textbf{SRCC} $\uparrow$ & \textbf{PCC} $\uparrow$ & \textbf{RMSE} $\downarrow$ & \textbf{RMAE} $\downarrow$ \\
\midrule
-- &       & \textbf{0.9509} & \textbf{0.9544} & \textbf{0.0495} & \textbf{0.1962} \\
a & m=1.0   & 0.9479 & 0.9518 & 0.0517 & 0.2011 \\
b & m=0.5   & 0.9508 & 0.9539 & 0.0512 & 0.1996 \\
c & rs     & 0.9396 & 0.9409 & 0.0559 & 0.2076 \\
d & mw     & 0.9238 & 0.9242 & 0.0851 & 0.2645 \\
\bottomrule
\end{tabular}
\end{table}

\begin{table}[t]
\centering
\scriptsize
\caption{\textbf{Comparisons with different captions.} 'Only Caption' indicates using captions alone for IC evaluation. 'Image \& Caption Concat' means concatenating the image features and text features before inputting them into the regression head. SRCC and PCC, larger $\uparrow$ is better. RMSE and RMAE, smaller $\downarrow$ is better.}
\label{tab:7}
\begin{tabular}{c|llll|llll}
\toprule
\multirow{2}{*}{\textbf{CapType}} & \multicolumn{4}{c|}{Only Caption} & \multicolumn{4}{c}{Image \& Caption Concat} \\ \cmidrule(l){2-9} 
 & \multicolumn{1}{c}{\textbf{SRCC}} & \multicolumn{1}{c}{\textbf{PCC}} & \multicolumn{1}{c}{\textbf{RMSE}} & \multicolumn{1}{c|}{\textbf{RMAE}} & \multicolumn{1}{c}{\textbf{SRCC}} & \multicolumn{1}{c}{\textbf{PCC}} & \multicolumn{1}{c}{\textbf{RMSE}} & \multicolumn{1}{c}{\textbf{RMAE}} \\ \midrule
CapI & 0.7095 & 0.7162 & 0.1084 & 0.2868 & 0.9456 & 0.9501 & 0.0535 & 0.2044 \\
CapII & 0.8120 & 0.8172 & 0.0903 & 0.2630 & 0.9469 & 0.9506 & 0.0534 & 0.2044 \\
CapIII & 0.8102 & 0.8151 & 0.0904 & 0.2626 & 0.9441 & 0.9484 & 0.0524 & \textbf{0.2015} \\
BLIP & \textbf{0.8260} & \textbf{0.8251} & \textbf{0.0888} & \textbf{0.2609} & \textbf{0.9476} & \textbf{0.9524} & \textbf{0.0524} & 0.2029 \\ \bottomrule
\end{tabular}
\end{table}

\subsection{Further Analysis}
\label{sec-exp-aaa}

\begin{figure}[t]
    \centering
    \includegraphics[width=1\linewidth]{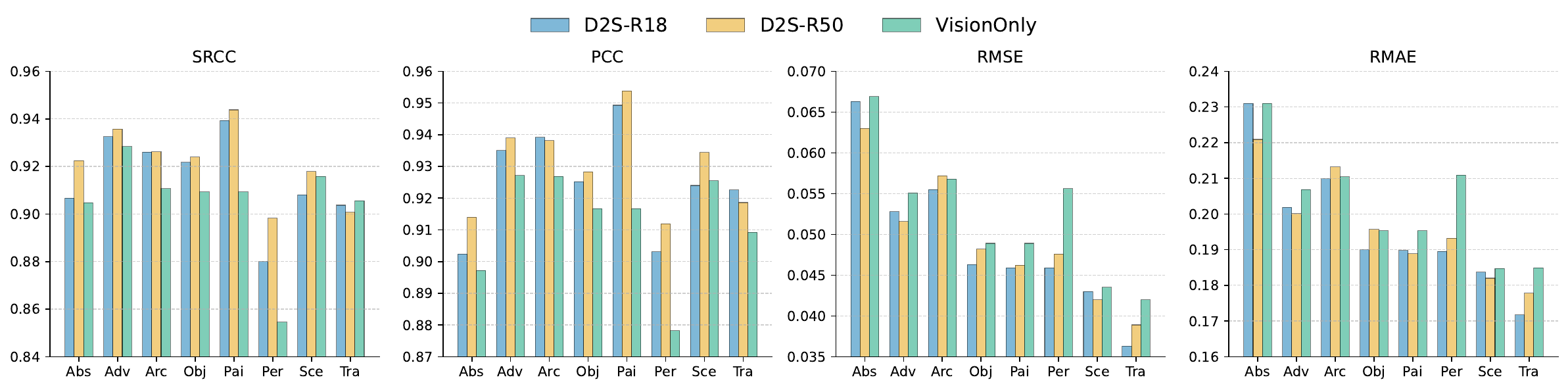}
    \caption{\textbf{Performance of semantic categories in IC9600.}}
    \label{ic9600_div}
\end{figure}

\textbf{Performance of semantic category in IC9600.} We assess how semantic alignment affects different semantic categories in IC9600 by evaluating D2S-R18, D2S-R50, and the VisionOnly variant (Figure \ref{ic9600_div}). D2S consistently surpasses VisionOnly across correlation (SRCC/PCC) and error metrics (RMSE/RMAE). The improvements are most pronounced in object-centric categories (\textit{Obj}, \textit{Per}) and abstract categories (\textit{Abs}, \textit{Pai}), where visual cues alone are insufficient. D2S-R50 further outperforms D2S-R18, especially in \textit{Arc} and \textit{Sce}, suggesting that larger visual backbones can better exploit semantic guidance. These results show that the proposed alignment strategy not only improves average accuracy but also stabilizes predictions across diverse semantic distributions.

\begin{figure}[t]
\centering
\begin{minipage}[t]{0.5\textwidth}
\centering
\includegraphics[width=1\linewidth]{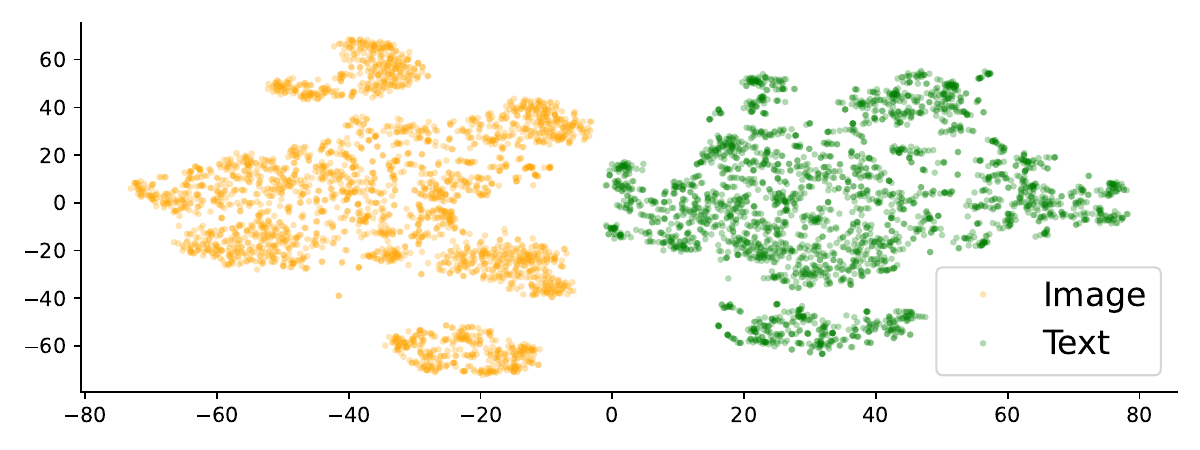}
\caption{\textit{t}-SNE before alignment}
\label{tsne_CLIP}
\end{minipage}\hfill
\begin{minipage}[t]{0.5\textwidth}
\centering
\includegraphics[width=1\linewidth]{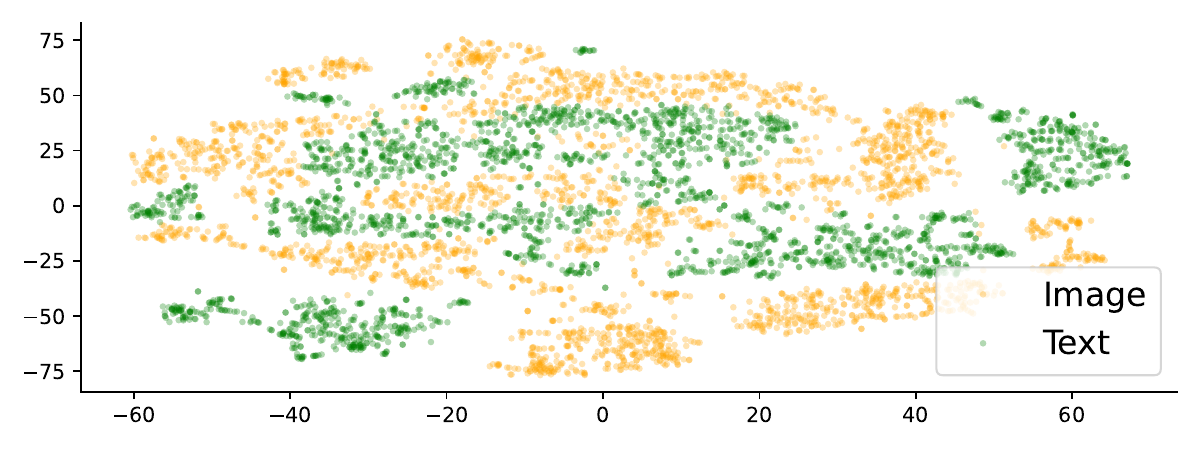}
\caption{\textit{t}-SNE after alignment}
\label{tsne_D2S}
\end{minipage}
\vspace{-10pt}
\end{figure}

\begin{figure}[t]
\centering
\begin{minipage}[t]{0.45\textwidth}
\centering
\includegraphics[width=1\linewidth]{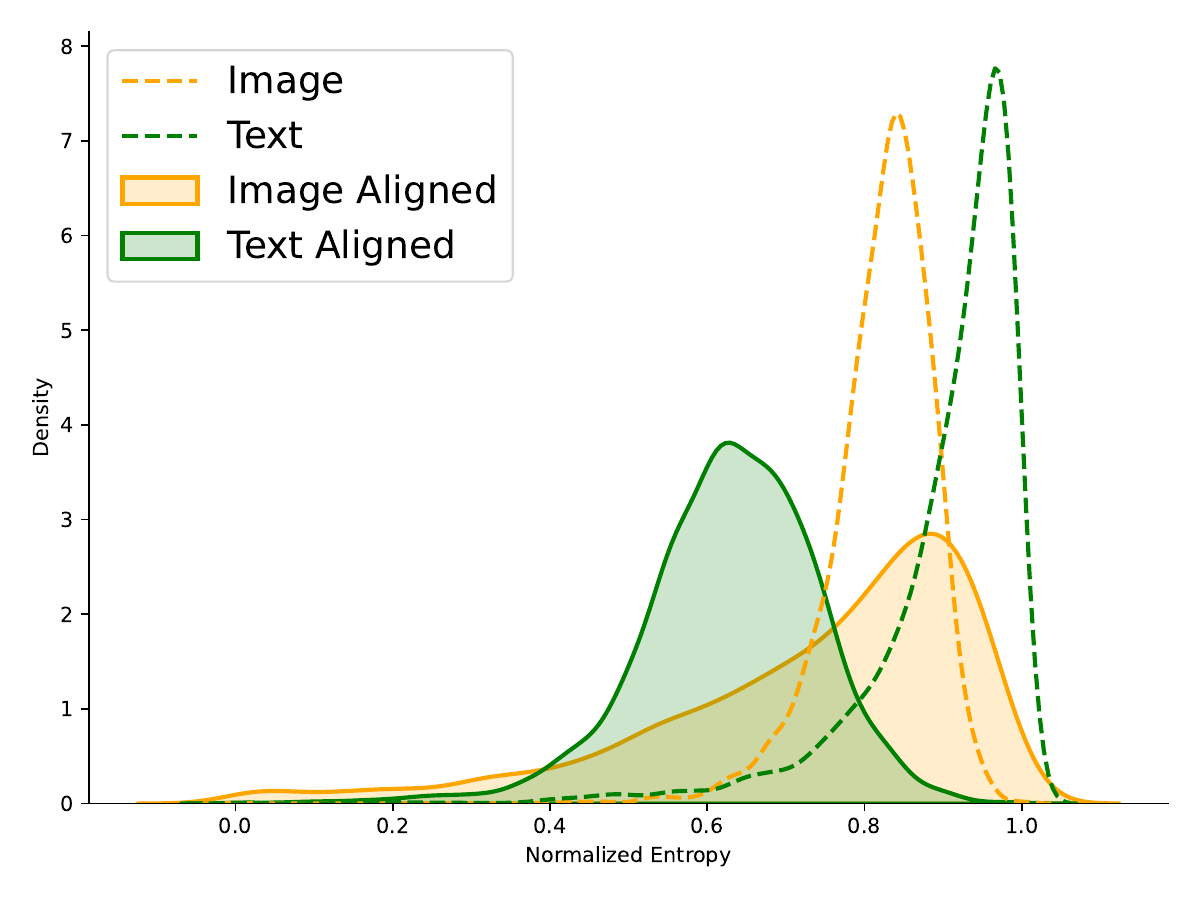}
\caption{\textbf{Entropy distribution alignment.} Visual and textual entropy distributions before (dashed) and after alignment (solid).}
\label{merge}
\end{minipage}\hfill
\begin{minipage}[t]{0.51\textwidth}
\centering
\includegraphics[width=1\linewidth]{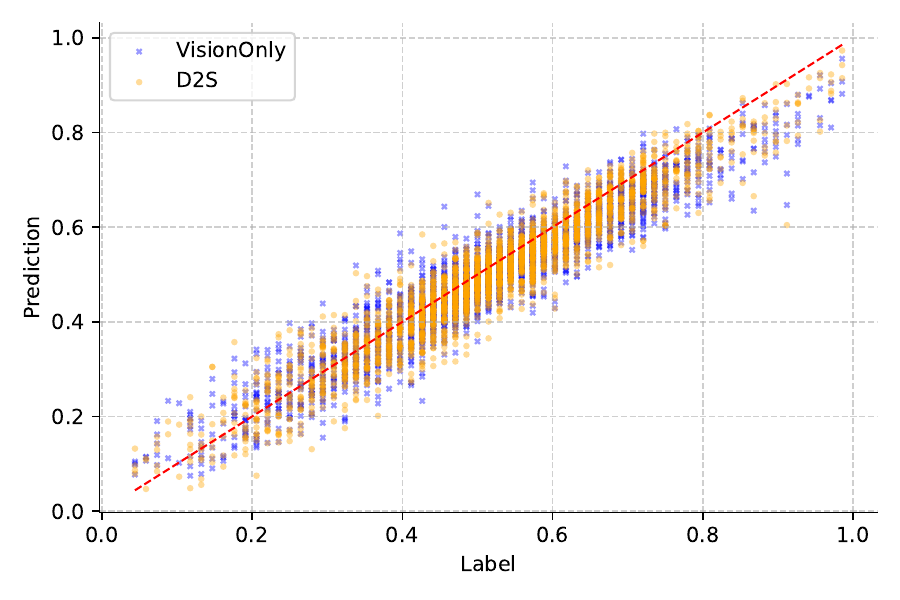}
\caption{\textbf{Prediction scatter plot.} Scatter plots of predicted scores versus ground-truth labels.}
\label{linear_scatter}
\end{minipage}
\end{figure}

\textbf{Visualization of modality alignment.} We visualize visual and textual embeddings using t-SNE before and after alignment. In Figure \ref{tsne_CLIP}, the two modalities form separate clusters. After alignment (Figure \ref{tsne_D2S}), the clusters become interleaved, indicating that the alignment module effectively bridges the semantic gap. This qualitative behavior matches the entropy statistics in Figure~\ref{merge}, where the overlap between visual and textual entropy distributions increases substantially after EAL.

\textbf{Prediction scatter and regression line.} Scatter plots in Figure \ref{linear_scatter} show that D2S predictions lie closer to the ideal regression line than the baseline. This indicates not only stronger correlation but also reduced systematic bias. The tighter dispersion reflects the stabilizing effect of semantic regularization, which improves local ranking consistency across complexity levels.

\begin{figure}[t]
    \centering
    \includegraphics[width=1\linewidth,height=140pt]{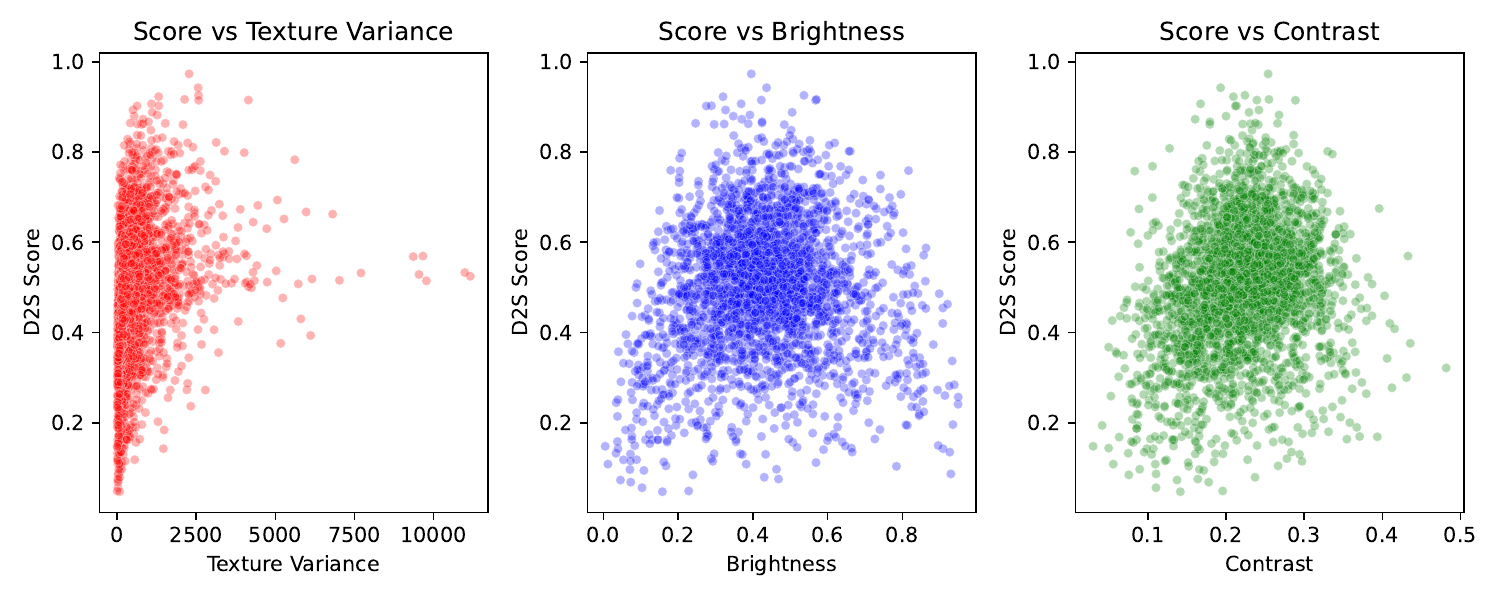}
    \caption{The correlation between the predicted scores of D2S and the low-level statistics, including texture variance, brightness, and contrast.}
    \label{scores_vs_others}
\end{figure}

\begin{figure}[t!]
\centering
\begin{minipage}[t]{0.47\textwidth}
\centering
\includegraphics[width=1\linewidth]{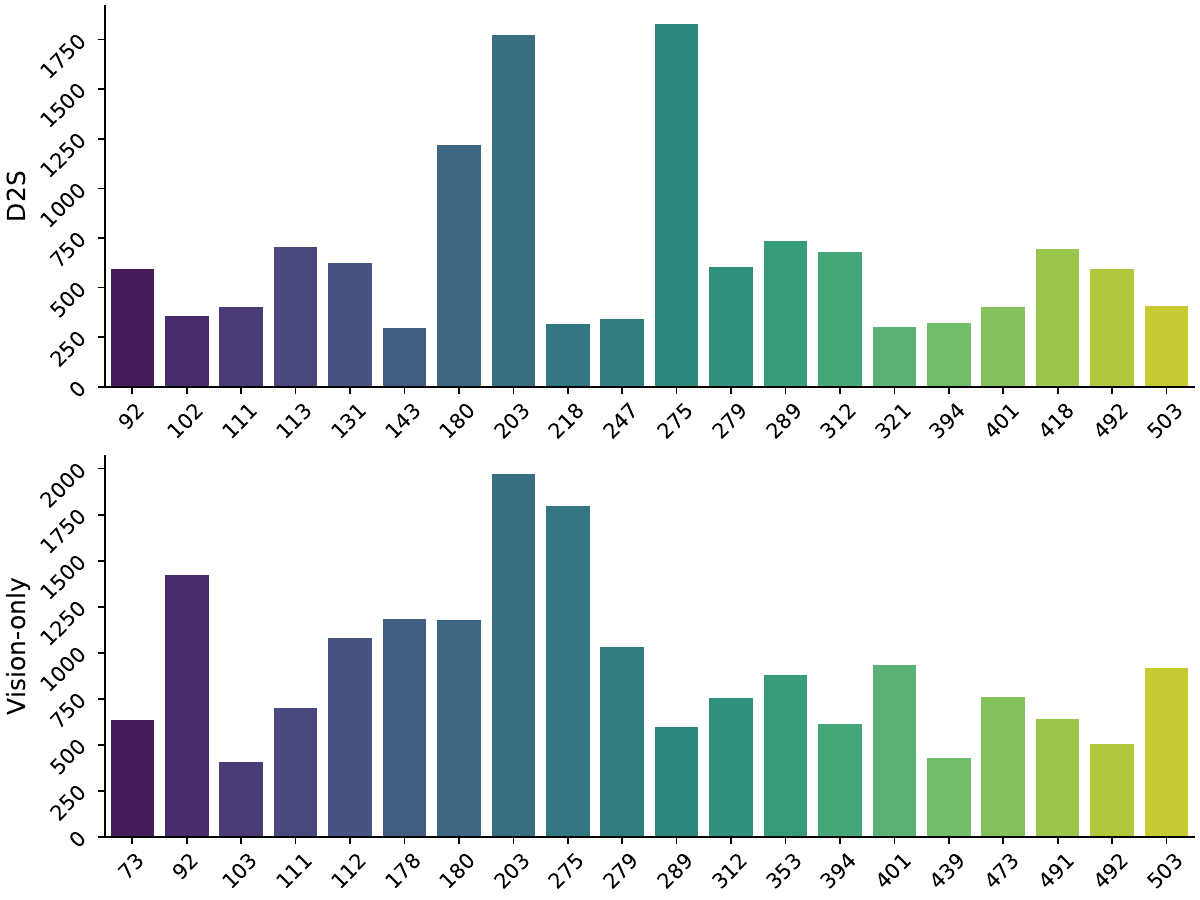}
\caption{\textbf{Channel utilization histogram.} Frequencies of the top-20 most correlated feature channels across the IC9600 test set. The above subgraph represents D2S, and the one below is Vision-only.} 
\label{top20ch}
\end{minipage}\hfill
\begin{minipage}[t]{0.47\textwidth}
\centering
\includegraphics[width=1\linewidth]{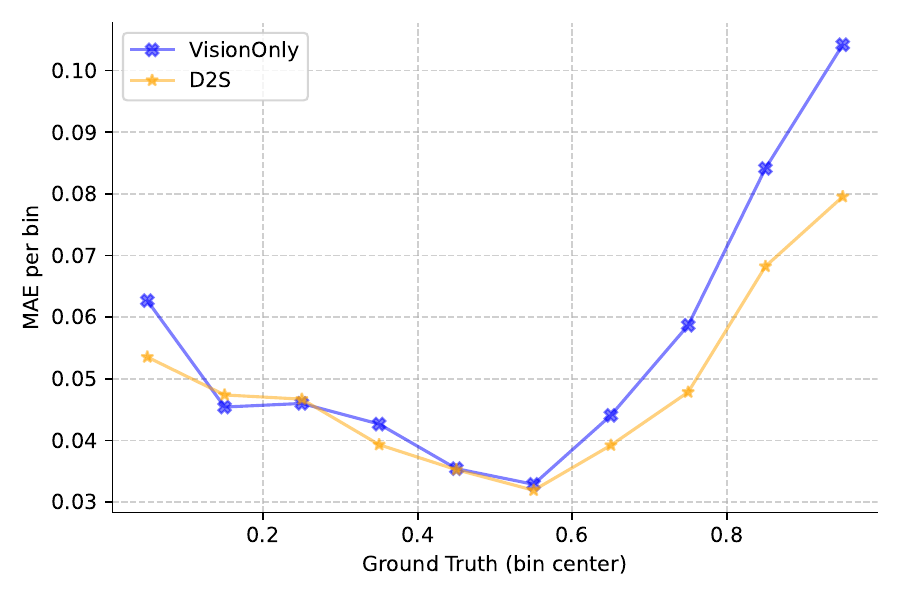}
\caption{\textbf{Binned MAE comparison.} Mean absolute error (MAE) across 10 uniformly spaced complexity bins.}
\label{mae_bins10}
\end{minipage}
\end{figure}

\textbf{What Does D2S Learn?} We analyze the top-20 frequently activated channels (Figure~\ref{top20ch}). D2S (top subplot) exhibits a concentrated and stable activation pattern, relying on a compact set of discriminative channels. These activations show no strong correlation with low-level image statistics such as brightness, contrast, or texture variance (Figure~\ref{scores_vs_others}), indicating that D2S does not rely on superficial cues but instead captures higher-level patterns relevant to complexity.

Compared with a purely visual baseline (bottom subplot), D2S suppresses noisy or weakly informative channels and promotes more structured channel usage. This structured sparsity is a direct effect of introducing semantic supervision and also explains the improved cross-dataset generalization.

\textbf{Binned Error Analysis.} We divide predicted scores into 10 uniform complexity bins (Figure \ref{mae_bins10}). D2S performs similarly to the visual-only baseline for simple images (complexity $<$ 0.55), but consistently outperforms it as complexity increases. The margin widens for high-complexity scenes, confirming that textual guidance is most beneficial when structural and semantic cues become important.

\begin{figure}[t!]
    \vspace{-10pt}
    \centering
    \includegraphics[width=1\linewidth]{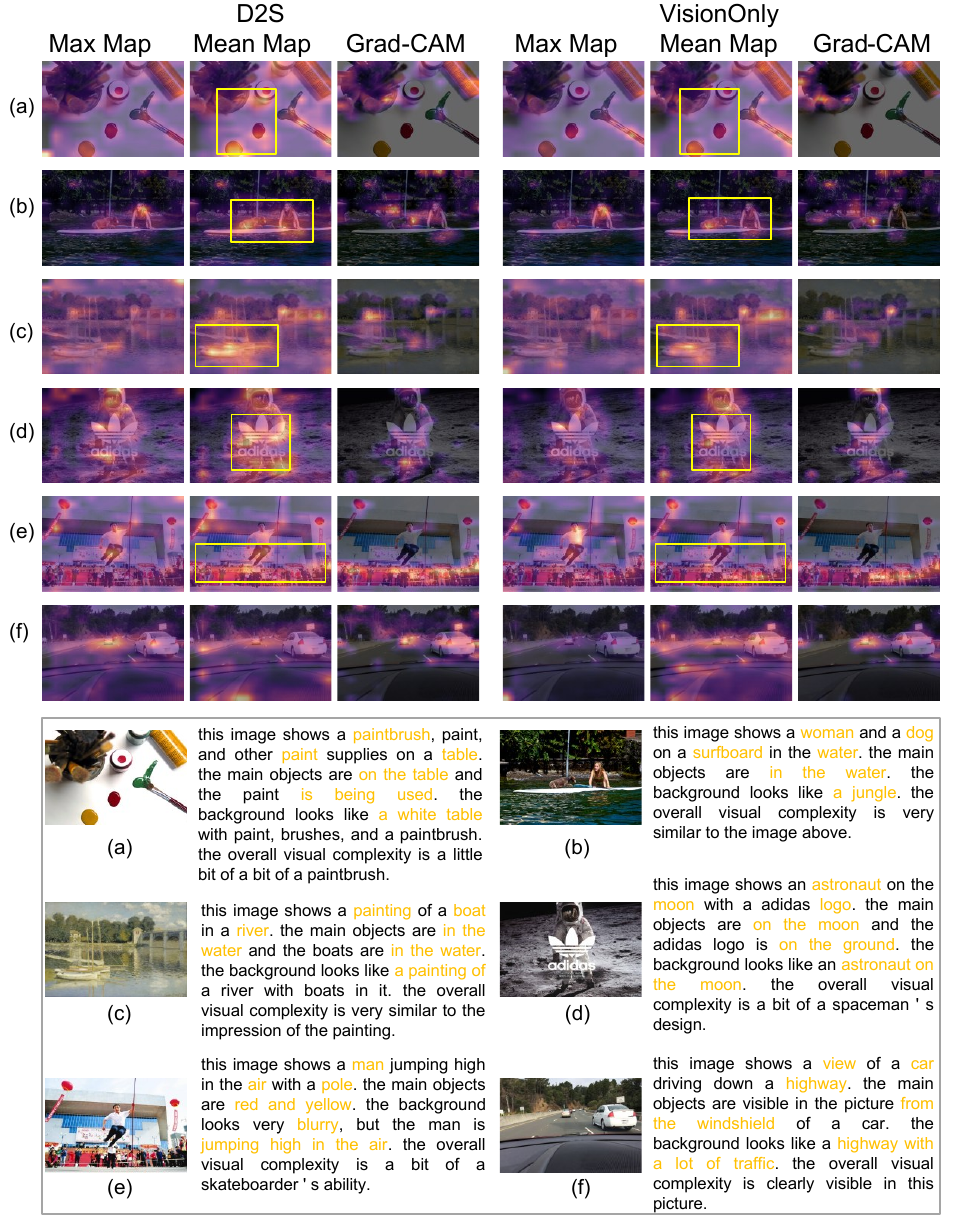}
    \caption{\textbf{Feature activation maps.} Visualization of channel activation maps (stage 4 in ResNet) using three approaches: maximum channel activation, mean channel activation, and Grad-CAM.}
    \label{act_map_compare}
    \vspace{-10pt}
\end{figure}

\textbf{Feature Activation Visualization.} We further analyze how D2S attends to image regions by comparing maximum activation, mean activation, and Grad-CAM visualizations (Figure \ref{act_map_compare}). Mean activation maps highlight the largest number of structural regions, offering a more global representation of complexity. Maximum activation maps capture salient object parts, while Grad-CAM produces sparser responses and tends to miss fine-grained details. Under semantic guidance, D2S exhibits concentrated activation near caption-related objects, indicating that textual cues steer the visual encoder toward semantically meaningful structures.

\section{Conclusion}
In this work, we introduced D2S, a text-guided framework for image complexity assessment that uses pre-trained vision-language models to provide semantic supervision during training. By combining feature alignment and entropy distribution alignment, D2S encourages the visual encoder to internalize complexity-relevant semantic structure. The theoretical analysis characterizes this process through enriched representational geometry and semantic constraints on the admissible predictor family. Experiments on IC9600 and related benchmarks validate these observations: D2S achieves state-of-the-art performance on ICA while maintaining a vision-only inference pipeline with no additional test-time multimodal cost, and it also shows competitive transferability to image quality assessment tasks. Overall, the results indicate that semantic guidance provides an effective and efficient way to regularize visual complexity modeling, with potential relevance to broader perceptual prediction problems.

\textbf{Limitations.} This work has several limitations. First, although D2S is reasonably robust to moderate caption perturbations, its effectiveness still depends on the quality and semantic relevance of the generated descriptions. Offline caption generation, the frozen text encoder, and the EMA-updated momentum model introduce additional preprocessing, training-time computation, and memory, although these components are absent from the inference pipeline. Second, while the improvements over strong visual baselines are consistent, their absolute magnitude is sometimes modest, especially when the visual backbone is already highly competitive. Third, the cross-dataset results on Savoias and VISC-C/I indicate that semantic guidance alone does not fully resolve annotation and distribution shifts. Fourth, the SST@500 analysis shows that the few-shot protocol is affected by long-tailed interval occupancy at the high-complexity end. Finally, the theoretical analysis in this work is intended as an idealized explanatory framework for the design of D2S rather than a universal characterization of all multimodal complexity models. Additional theoretical notes on entropy non-universality and semantic-quality dependence are provided in \ref{supp-Practical Notes and Limitations}.

Future work will focus on stronger transfer mechanisms, more reliable semantic supervision, and more balanced sampling strategies for high-complexity regimes.

\section{Acknowledgements}
This work would not have been possible without financial support from the Science and Technology Innovation Team of Shaanxi Innovation Capability Support Plan (2020TD005), Xi'an Scientists \& Engineers Workforce Building Project (2024JH-KGDW-0112), Key Project of the Natural Science Basic Research Program of Shaanxi Province (2025SYS-SYSZD-049). The authors are also grateful to the editors and anonymous reviewers for their sound and valuable suggestions on the manuscript.

\section{Declaration of generative AI and AI-assisted technologies in the manuscript preparation process}
During the preparation of this work the authors used ChatGPT-5 for translation and language polishing. After using this tool/service, the authors reviewed and edited the content as needed and take full responsibility for the content of the published article.

\bibliographystyle{elsarticle-num}
\bibliography{refs}

\newpage
\setcounter{page}{1}
\setcounter{table}{0}
\setcounter{figure}{0}
\appendix
{\centering \Large Describe-to-Score: A Text-Guided Framework for Image Complexity Assessment \\ \par}

$\quad$

{\centering Supplementary material \par}

\section{Appendix}
\label{app1}

\subsection{Theoretical Foundations and Empirical Analysis of Multimodal Fusion}

\subsubsection{Justification of Proposition 1}
\label{supp-Justification of Proposition 1}

We first define the entropy of visual and semantic feature distributions. Assume the visual encoder outputs a feature distribution $\{p_{k}^{(v)}\}_{k=1}^{K}$, and the text encoder outputs a feature distribution $\{p_{t}^{(s)}\}_{t=1}^{T}$. Their corresponding entropies are defined as:
\begin{equation}
    H_{v}^{F}(I) = - \sum_{k=1}^{K} p_{k}^{(v)} \log p_{k}^{(v)}, \qquad
    H_{s}^{F}(S) = - \sum_{t=1}^{T} p_{t}^{(s)} \log p_{t}^{(s)}
    \label{eq-entropy}
\end{equation}

Here, $H_{v}^{F}(I)$ measures the diversity of the visual space (e.g., texture, color, structure), and $H_{s}^{F}(S)$ measures the diversity of the semantic space (e.g., object categories, relations, actions). To analyze multimodal fusion, we consider a linear entropy model:
\begin{equation}
    H^{F}(I) = \alpha\, H^{F}_{v}(I) + \beta\, H^{F}_{s}(S)
\end{equation}
where $\alpha, \beta > 0$ are weighting coefficients.

This expression is used as a tractable approximation for analyzing how visual and semantic entropies jointly contribute to the fused representation. It should not be interpreted as a universal identity or as a direct definition of perceived image complexity. In particular, $H^{F}(I)>H_v^{F}(I)$ is not guaranteed for arbitrary $\alpha$ and $\beta$. Rather, the model captures the empirical tendency that complexity-relevant textual semantics can introduce additional representational diversity when the semantic branch provides complementary cues.

In our trained model, we estimate $\alpha$ and $\beta$ by fitting the linear relation to the fused entropy measured from model activations. The fitted coefficients are positive, and the empirical results in \ref{supp-alpha-beta} show that this linear form provides a reasonable approximation of the observed entropy trend. Therefore, Proposition~1 should be understood as an empirical modeling principle that motivates entropy-based alignment, rather than as a strict monotonicity claim.

Empirical statistics further show that for the vast majority of samples, the fused entropy exceeds the visual entropy. This supports the use of entropy distribution alignment as a training-time regularizer that encourages cross-modal consistency. However, higher fused entropy alone does not guarantee better prediction accuracy; its benefit depends on whether the added semantic diversity corresponds to complexity-relevant factors such as object count, object categories, spatial layout, and scene organization.

\subsubsection{Proof of Proposition 2}
\label{supp-proof2}

\textbf{Proof.} By definition, $\mathcal{F}_{\epsilon}$ contains only predictors in $\mathcal{F}_{v}$ that satisfy the semantic alignment constraint. Therefore,
\begin{equation}
\mathcal{F}_{\epsilon} \subseteq \mathcal{F}_{v}.
\end{equation}

For any fixed realization of the Rademacher variables $\sigma$, the supremum over a subset cannot exceed the supremum over the original set:
\begin{equation}
\sup_{f\in\mathcal{F}_{\epsilon}} \frac{1}{n} \sum_{i=1}^{n} \sigma_i f(x_i) \leq \sup_{f\in\mathcal{F}_{v}} \frac{1}{n} \sum_{i=1}^{n} \sigma_i f(x_i).
\end{equation}

Taking expectation over $\sigma$ on both sides gives
\begin{equation}
\widehat{\mathcal{R}}_{\mathcal{D}_{n}}
\left(
\mathcal{F}_{\epsilon}
\right)
\leq
\widehat{\mathcal{R}}_{\mathcal{D}_{n}}
\left(
\mathcal{F}_{v}
\right).
\end{equation}

\subsubsection{Empirical Estimation of Fusion Coefficients}
\label{supp-alpha-beta}

To validate the linear fusion model used in Proposition~1, we perform a regression analysis to estimate the coefficients $\alpha$ and $\beta$ directly from the fused entropy produced by our trained D2S model. Given $N$ training samples, we fit
\begin{equation}
    H^{F}(I) \approx \alpha\, H_{v}^{F}(I) + \beta\, H_{s}^{F}(S)
\end{equation}
using least squares. The resulting fits are:
\[
\text{(1) Visual-only:}\quad
\alpha_{\text{vis}} = 0.2486,\;
\beta_{\text{vis}} \approx 0,\;
R^{2}=0.96.
\]
\[
\text{(2) Vision–text fusion:}\quad
\alpha_{\text{mul}} = 0.4490,\;
\beta_{\text{mul}} = 1.1045,\;
R^{2}=0.58.
\]
The positive coefficients and the moderate-to-high fitting quality indicate that the linear model captures the dominant trend of entropy fusion in practice, although it should not be interpreted as an exact identity. Notably, $\beta_{\text{mul}}>1$ suggests that semantic entropy contributes substantially to the fused entropy, consistent with the increased representational diversity observed in our experiments.

\subsubsection{Synthetic Validation of Entropy Fusion}
\label{supp-synth}

To further validate Proposition~1 under controlled conditions, we conduct a synthetic experiment where the assumptions of the linear model hold exactly. We construct a high-rank Gaussian visual distribution with covariance $\Sigma_v$, a low-rank semantic distribution with covariance $\Sigma_s$,
and a fused distribution $Z_f = \alpha Z_v + \beta Z_s$. Entropy for multivariate Gaussian is given by
\begin{equation}
    H(Z) = \frac{1}{2}\log\big((2\pi e)^{d}\det(\Sigma)\big)
\end{equation}
By varying $\alpha$ and $\beta$ over a grid, we observe that $H(Z_f)$ increases monotonically with $\beta$, the increase is largest when $\Sigma_s$ contributes new orthogonal directions to $\Sigma_v$, the synthetic results match the empirical trends seen in real models.

This controlled experiment confirms that when semantic features add complementary variance directions, the fused entropy indeed grows, supporting the intuition behind Proposition~1.

\subsubsection{Representation Geometry under Semantic Alignment}
\label{supp:id-geometry}

We further examine how semantic alignment affects the geometry of the learned representations. Given a feature matrix $Z$, we estimate its intrinsic dimension using the participation-ratio estimator, defined as

\begin{equation}
\operatorname{ID}_{\mathrm{PR}}(Z)
=
\frac{\left(\operatorname{tr}(\mathbf{C}_{Z})\right)^2}
{\operatorname{tr}\left(\mathbf{C}_{Z}^{2}\right)}
=
\frac{\left(\sum_{k}\lambda_{k}\right)^2}
{\sum_{k}\lambda_{k}^{2}},
\end{equation}

where $\mathbf{C}_{Z}$ denotes the empirical covariance matrix of $Z$, and $\{\lambda_k\}$ denotes its eigenvalue spectrum. The participation ratio measures the effective number of variance-bearing directions in a representation. It therefore characterizes feature geometry and should not be interpreted as an empirical estimate of the capacity of the corresponding predictor family.

Let $Z_v^{\mathrm{pre}}$ and $Z_v^{\mathrm{align}}$ denote the visual representations before and after semantic alignment, respectively. Similarly, let $Z_j^{\mathrm{pre}}$ and $Z_j^{\mathrm{align}}$ denote the corresponding joint vision--text representations. The textual representation is denoted by $Z_s$. All intrinsic dimensions are computed on the same set of samples using an identical feature-processing and estimation procedure.

\begin{table}[t]
\centering
\caption{Intrinsic dimensions of the visual, joint vision--text, and textual representations. All values are computed using the participation-ratio estimator on the same set of samples. The text encoder remains frozen during 
training.}
\label{tab:id-analysis}
\begin{tabular}{lcc}
\toprule
Representation & Before alignment & After alignment \\
\midrule
Visual
& $1.748$ & $14.150$ \\
Vision--Text
& $4.910$ & $28.407$ \\
Textual
& $50.235$ & $50.235$ \\
\bottomrule
\end{tabular}
\end{table}

As shown in Table~\ref{tab:id-analysis}, the intrinsic dimension of the visual representation increases from $1.748$ before alignment to $14.150$ after alignment. The intrinsic dimension of the joint vision--text representation similarly increases from $4.910$ to $28.407$, whereas the frozen textual representation remains at $50.235$.

Before semantic alignment, both the visual and joint representations exhibit strongly concentrated covariance spectra, with most of their variation captured by only a small number of effective directions. In particular, the low intrinsic dimension of $Z_v^{\mathrm{pre}}$ indicates a highly anisotropic visual representation under direct scalar regression supervision. Semantic alignment broadens the effective spectrum of the visual representation, allowing the visual branch to preserve a wider range of task-relevant variation directions. These directions may encode complementary factors associated with object composition, spatial organization, scene structure, and foreground--background relationships.

The increase in the intrinsic dimension of the joint representation indicates that semantic alignment also preserves a richer set of effective cross-modal variations. The resulting dimensional ordering characterizes the relative covariance geometry of the visual, joint, and textual representations. The aligned visual and joint representations occupy broader effective subspaces than their pre-alignment counterparts while remaining geometrically distinct from the frozen textual representation. Because these representations are distinct variables rather than the input and output of an explicitly defined compression operator, the ordering does not establish a compression mapping between the textual and visual spaces.

The observed dimensional expansion is compatible with Proposition~2. Intrinsic dimension characterizes the covariance geometry of feature samples, whereas empirical Rademacher complexity characterizes the capacity of a family of prediction functions. Consequently, a representation with a broader effective spectrum can coexist with a more restricted family of semantically admissible predictors. In D2S, semantic alignment increases task-relevant representational degrees of freedom while excluding predictors that are inconsistent with caption-derived semantic structure.

\subsubsection{Practical Notes and Limitations}\label{supp-Practical Notes and Limitations}

We highlight two practical considerations.

(1) \textit{Non-universality of entropy monotonicity.}
The inequality $H^{F}(I)>H_v^{F}(I)$ holds for most samples in our trained model but is not mathematically universal. Caption noise, low-quality semantic descriptions, or very small $\beta$ may lead to exceptions. This is consistent with Proposition~1, which is framed as an approximate modeling principle rather than a strict entropy theorem.

(2) \textit{Dependence on semantic quality.}
If the caption generator or text encoder produces low-quality or irrelevant semantic information, the benefit of entropy enrichment and semantic alignment may diminish. Our robustness experiments in Table \ref{tab:badcap} show that D2S remains stable under moderate caption perturbations but degrades when the text input becomes meaningless.

These considerations define the empirical scope under which Proposition~1 and Proposition~2 are most informative: semantic guidance is useful when the textual branch provides complexity-relevant structure, and its role is to regularize the visual hypothesis space rather than to guarantee universal improvement.

\subsection{EAL Algorithm}
\label{supp-algorithm}
The complete training procedure of entropy distribution alignment, including buffer initialization, FIFO update, momentum refresh, and loss computation, is summarized in Algorithm~\ref{algorithm1}.

\begin{algorithm}[H]
\label{algorithm1}
\caption{Entropy Distribution Alignment}
\KwIn{Image $I$, Caption $S$, D2S $f_{\theta}$, MoM $f_{\xi}$, Momentum $m$, FIFO Buffers $B_v, B_s$, Buffer size $M$, Refresh step $r$}
\do{\textbf{Before training:} Create $f_{\theta}$, copy as $f_{\xi}$; Create $B_v, B_s$ with $M$.}

\For{each mini-batch}{
    Extract features $z_v$ and $z_s$ using $f_\xi$ via Eq.(\ref{eq-feat})\;
    Compute entropies $H_v$ and $H_s$ via Eq.(\ref{eq-entropy})\;
    Enqueue new entropy values into $B_v$ and $B_s$ and dequeue the oldest entries\;
    \If{$|B_v| \ge M/2$ and $|B_s| \ge M/2$}{
        Compute $\mathcal{L}_{\mathrm{eal}}$ from $B_v$ and $B_s$\;
    } \Else {
        $\mathcal{L}_{\mathrm{eal}} = 0$
    }
    Update $f_{\theta}$ via gradient\;
    Update $f_{\xi}$ via $\xi \gets m \xi + (1-m)\theta$\;
    Refresh $r$ old entries via updated $f_{\xi}$\;
}
\end{algorithm}

\begin{table}[t]
\centering
\scriptsize
\setlength{\tabcolsep}{10pt}
\caption{\textbf{Full-data training results on Savoias, Nagle4K, and VISC-C.} Each dataset is split into training and testing sets with an 8:2 ratio, and all training hyperparameters follow those used in the main paper.}
\label{tab:fulltrain}
\begin{tabular}{lcccccc}
\toprule
\multirow{2}{*}{\textbf{Method}} & \multicolumn{2}{c}{\textbf{Savoias}} & \multicolumn{2}{c}{\textbf{Nagle4K}} & \multicolumn{2}{c}{\textbf{VISC-C}} \\
\cmidrule(lr){2-3} \cmidrule(lr){4-5} \cmidrule(lr){6-7}
 &  \textbf{SRCC} $\uparrow$ & \textbf{PCC} $\uparrow$ &  \textbf{SRCC} $\uparrow$ & \textbf{PCC} $\uparrow$ &  \textbf{SRCC} $\uparrow$ & \textbf{PCC} $\uparrow$ \\
\midrule
DBCNN       &  0.768 & 0.770 &  0.745 & 0.732 &  0.779 & 0.783 \\
NIMA        &  0.781 & 0.771 &  0.756 & 0.741 &  \underline{0.810} & 0.803 \\
HyperIQA    &  0.801 & 0.798 &  0.772 & 0.758 &  0.734 & 0.739 \\
CLIPIQA     &  0.779 & 0.794 &  0.785 & 0.774 &  0.781 & 0.796 \\
TOPIQ       &  0.838 & 0.832 &  0.804 & 0.792 &  0.803 & \underline{0.811} \\
ICNet       &  0.845 & 0.849 &  0.815 & 0.804 &  0.789 & 0.790 \\
ICCORN      &  0.852 & 0.856 &  0.818 & 0.806 &  0.796 & 0.793 \\
D2S-R18     &  \underline{0.861} & \underline{0.864} &  \underline{0.823} & \underline{0.813} &  0.801 & 0.802 \\
D2S-R50     &  \textbf{0.875} & \textbf{0.876} &  \textbf{0.834} & \textbf{0.825} &  \textbf{0.822} & \textbf{0.820} \\
\midrule[0.8pt]
 &  \textbf{RMSE} $\downarrow$ & \textbf{RMAE} $\downarrow$ &  \textbf{RMSE} $\downarrow$ & \textbf{RMAE} $\downarrow$ &  \textbf{RMSE} $\downarrow$ & \textbf{RMAE} $\downarrow$ \\
\midrule
DBCNN       &  0.147 & 0.355 &  0.121 & 0.310 &  0.086 & 0.351 \\
NIMA        &  0.210 & 0.368 &  0.134 & 0.315 &  0.125 & 0.362 \\
HyperIQA    &  0.293 & 0.392 &  0.145 & 0.327 &  0.181 & 0.388 \\
CLIPIQA     &  0.171 & 0.348 &  0.112 & 0.302 &  0.122 & 0.348 \\
TOPIQ       &  0.123 & 0.330 &  0.101 & 0.290 &  0.079 & 0.340 \\
ICNet       &  0.121 & 0.325 &  0.097 & 0.278 &  0.123 & 0.349 \\
ICCORN      &  \underline{0.119} & 0.324 &  \textbf{0.088} & \underline{0.265} &  0.119 & 0.342 \\
D2S-R18     &  0.128 & \underline{0.321} &  0.092 & 0.268 &  \textbf{0.115} & \textbf{0.328} \\
D2S-R50     &  \textbf{0.114} & \textbf{0.301} &  \underline{0.089} & \textbf{0.265} &  \underline{0.116} & \underline{0.332} \\
\bottomrule
\end{tabular}
\end{table}

\subsection{Full-Data Training on Other ICA Benchmarks.}
\label{supp-full data}
We train D2S from scratch on Savoias, Nagle4K, and VISC-C. Table~\ref{tab:fulltrain} shows that D2S-R50 consistently achieves the highest correlations and lowest errors on all three datasets, confirming that D2S generalizes well even without IC9600 pre-training.

\subsection{Related Works}
\label{supp-related works}

\subsubsection{Image Complexity Assessment.}
\textbf{Statistical features.} Early methods for image complexity assessment primarily relied on statistical features or low-level visual indicators, such as entropy~\citep{28}, symmetry~\citep{23,34}, spatial layout~\citep{25}, and compressibility~\citep{29,30}. These approaches are easy to implement and highly interpretable, but they suffer from clear limitations. Specifically, they mainly capture local structures and texture details, while being sensitive to noise and image resolution.

\textbf{Deep learning models.} With the advent of deep learning, several approaches attempted to directly model image complexity using learning-based frameworks~\citep{Nagle4k,saraee2020visual,34}. For instance, ICNet~\citep{feng2022ic9600} improves complexity regression by combining multi-scale inputs with convolutional features, while ICCORN~\citep{guo2023iccorn} employs a larger backbone and integrates ordinal regression constraints to enhance perceptual modeling. Celona et al.~\citep{celona2024use} further introduced ViTs for complexity assessment, highlighting the potential of deep features in complexity modeling. Nevertheless, these models largely focus on low-level image cues related to complexity, overlooking the human tendency to rely on high-level semantic information when making judgments.

\textbf{High-level semantics.} To our knowledge, Shen et al.~\citep{shen2024simplicity} were the first to introduce high-level semantics into this field. They quantified the number of segments and categories in an image using SAM~\citep{kirillov2023sam} and FC-CLIP~\citep{yu2023fcclip}, respectively, and employed a linear regression model to predict complexity scores. This improved interpretability, but yielded suboptimal performance. Li et al.~\citep{li2025micm} argued that implicit motion in image objects could benefit ICA, leveraging VLMs to generate simple captions that served as prompts to convert static images into dynamic videos. By fusing video, image, and text branches, they achieved the best performance to date on IC9600, but at the cost of $\sim$11B parameters and substantial computational resources.

\textbf{Our approach.} In contrast, our approach extracts high-level semantic information from VLMs through carefully designed prompt templates, using it only to guide the image branch during training. \textbf{At inference, our model requires only the visual input.}

\subsubsection{Vision-Language Modeling.}
In recent years, visual-language models (VLMs) have achieved remarkable progress in tasks such as image captioning~\citep{li2022blip}, cross-modal retrieval~\citep{wang2025cross}, and visual question answering~\citep{kuang2025natural}, with representative models including CLIP, BLIP, and ALIGN~\citep{align}. By aligning vision and language representations, these models effectively capture the semantic information of objects, relationships, and scenes. Prior studies show that language descriptions provide complementary information beyond low-level features, enhancing a model’s ability to understand and quantify image content. However, research on visual-language alignment mechanisms for image complexity modeling remains limited. Current complexity assessment methods have not fully exploited semantic information to improve cross-domain generalization. This gap motivates our Describe-to-Score framework, which achieves unified modeling of low-level vision and high-level semantics through visual-language fusion, thereby enhancing both the accuracy and generalization of complexity assessment.

\subsection{Image Caption examples}
\label{supp-capexp}
\begin{figure}[t]
    \centering
    \includegraphics[width=1\linewidth]{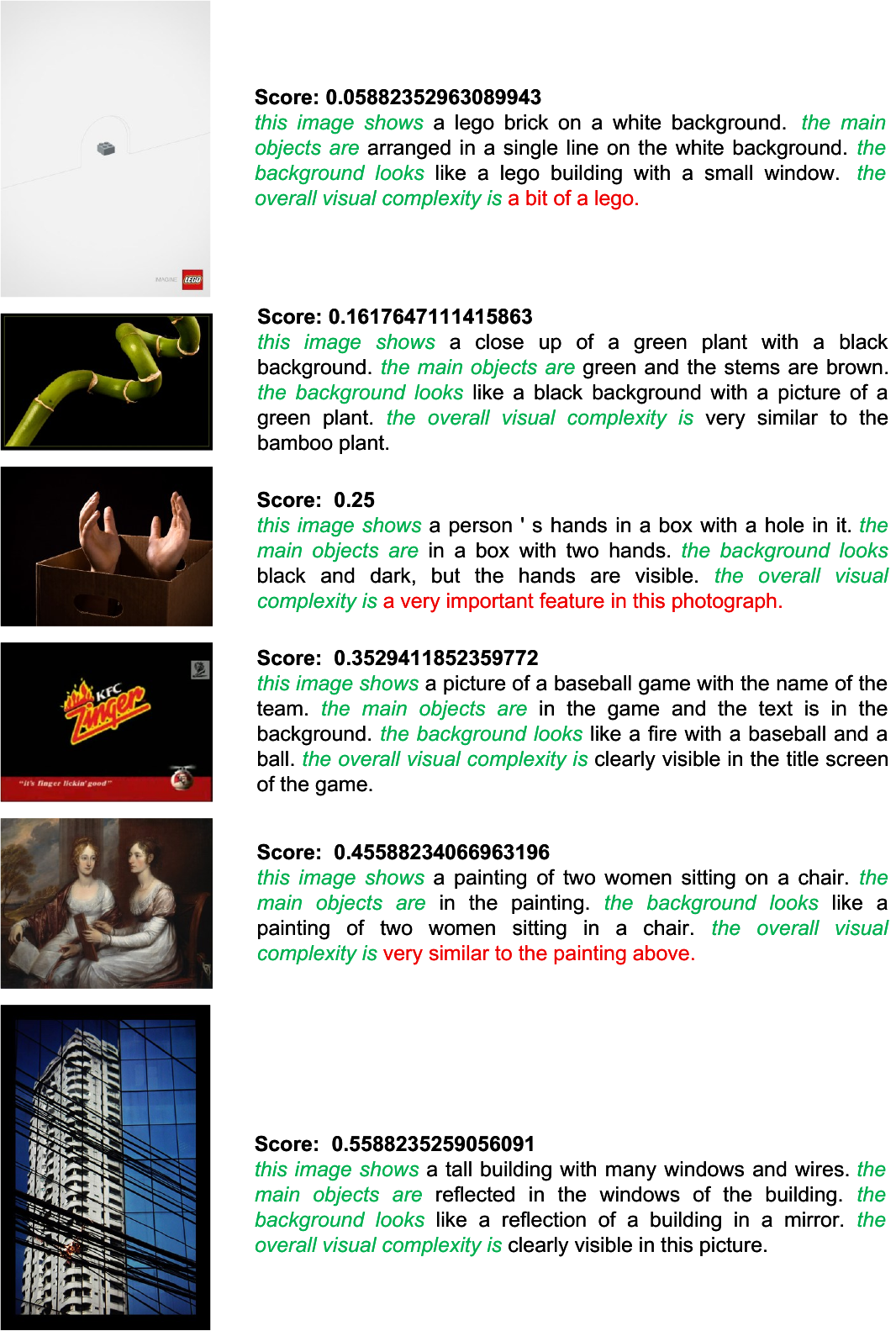}
    \caption{Image-caption pair examples (part 1).}
    \label{caption_exp1}
\end{figure}

\begin{figure}[t]
    \centering
    \includegraphics[width=1\linewidth]{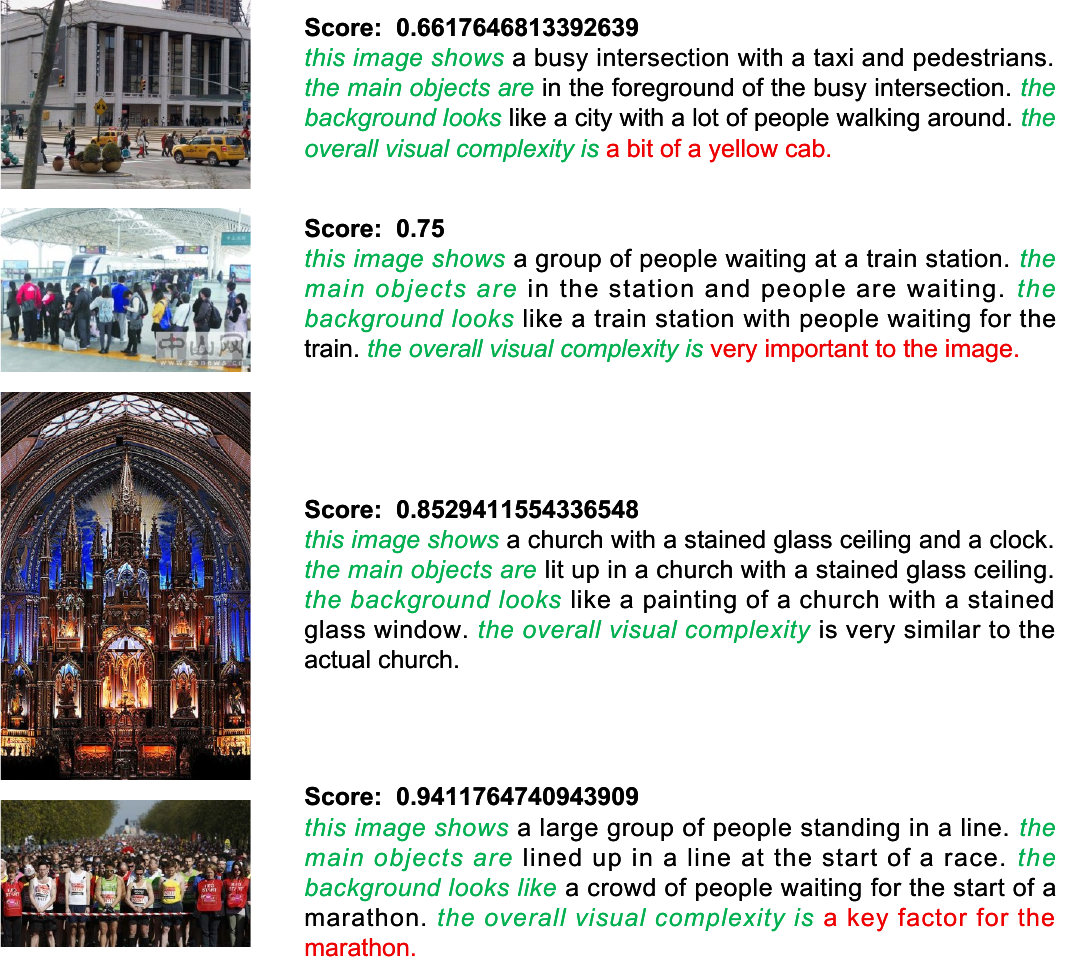}
    \caption{Image-caption pair examples (part 2).}
    \label{caption_exp2}
\end{figure}

In Figure \ref{caption_exp1} and \ref{caption_exp2}, we present some examples of captions. We selected one image-caption pair from each of the ten score ranges. The prompt template text is marked in green, while the generated text is marked in black or red. We observe that the responses to the last prompt are more error-prone because the prompt asks for an abstract global judgment that is difficult for the captioner to ground in visible image evidence.

\subsection{Discussion}
\label{supp-discuss}
\textbf{Does ICA really require high-level semantic information?}

A fundamental question in this work is whether high-level semantic information is necessary for accurate image complexity assessment. Low-level visual cues such as texture, color, and edge density provide a baseline measure of complexity, but human perception may also rely on semantic content, which can potentially improve prediction accuracy. 

\begin{table}[t]
\centering
\small
\setlength{\tabcolsep}{8pt}
\caption{Comparison of complexity predictions for different modalities. All the results are derived from the aforementioned table (rounded to three decimal places).}
\label{tab:8}
\begin{tabular}{lc|cccc}
\toprule
\textbf{Method} & \textbf{Modal} & \textbf{SRCC} $\uparrow$ & \textbf{PCC} $\uparrow$ & \textbf{RMSE} $\downarrow$ & \textbf{RMAE} $\downarrow$ \\ \midrule
CapI & \multirow{2}{*}{Text-only} & 0.710 & 0.716 & 0.108 & 0.287 \\
BLIP &  & 0.826 & 0.825 & 0.089 & 0.261 \\ \midrule
CLICv2 & \multirow{4}{*}{Vision-only} & 0.879 & 0.870 & - & - \\
HyperIQA &  & 0.935 & 0.935 & 0.067 & 0.229 \\
ICNet &  & 0.945 & 0.947 & 0.058 & 0.216 \\
ICCORN &  & 0.946 & 0.949 & 0.053 & 0.209 \\ \midrule
MICM & \multirow{3}{*}{Vision-Text fusion} & 0.943 & 0.953 & 0.060 & - \\
D2S-R18 &  & \underline{0.951} & \underline{0.954} & \underline{0.050} & \underline{0.196} \\
D2S-R50 &  & \textbf{0.954} & \textbf{0.958} & \textbf{0.050} & \textbf{0.196} \\ \bottomrule
\end{tabular}
\end{table}

Table~\ref{tab:8} compares the performance of multiple methods across three modalities: Text-only (CapI, BLIP), Vision-only (CLICv2, HyperIQA, ICNet, ICCORN), and Vision-Text fusion (MICM, D2S R18/R50). Text-only methods  rely purely on image captions. CapI achieves PCC 0.716 and BLIP improves to 0.825, indicating that textual cues capture part of the complexity information but are insufficient for high-precision prediction. Vision-only methods leverage visual features, with unsupervised methods like CLICv2 (PCC 0.870) learning content-invariant complexity representations. Supervised visual models reach PCC 0.949, highlighting that low-level visual features dominate complexity perception and provide strong baseline performance. Vision-Text fusion methods integrate both visual and semantic cues. MICM achieves PCC 0.953, and D2S-R50 reaches 0.958, with corresponding RMSE and RMAE showing the lowest prediction errors. Compared to vision-only baselines, this demonstrates a clear improvement, suggesting that semantic guidance provides measurable benefits in complexity assessment.

\textbf{Observations from these results.} Textual information alone is limited. While caption-based models partially capture complexity cues (PCC 0.716–0.825), they cannot match vision-based performance. Vision dominates baseline performance. Supervised vision-only models already achieve high PCC (0.947–0.949), confirming that low-level visual features are sufficient for most complexity signals. Semantic cues refine predictions. Vision-text fusion models consistently outperform vision-only methods, albeit the improvement is smaller than the jump from text-only to vision-only. This indicates that high-level semantics are not strictly necessary but provide a subtle enhancement, particularly for fine-grained correlation with human perception.

Therefore, the progression from Text-only to Vision-only to Vision-Text fusion highlights that while low-level visual features are the primary driver of complexity prediction, high-level semantic information can further refine accuracy. D2S exemplifies this synergy, leveraging semantic cues to enhance performance without replacing the foundational value of visual representation.

\end{document}